\documentclass[letterpaper]{article} 
\usepackage{aaai24}  
\usepackage{times}  
\usepackage{helvet}  
\usepackage{courier}  
\usepackage[hyphens]{url}  
\usepackage{graphicx} 
\usepackage{natbib}  
\usepackage{caption} 
\usepackage{algorithm}
\usepackage{algorithmic}
\usepackage{amsmath}
\usepackage{amssymb}
\usepackage{xcolor}
\usepackage{booktabs}
\usepackage{multirow}
\usepackage{tablefootnote}
\usepackage[shortlabels]{enumitem}
\usepackage{tensor}
\usepackage{gensymb}

\usepackage[capitalize]{cleveref}

\usepackage{newfloat}
\usepackage{listings}
\DeclareCaptionStyle{ruled}{labelfont=normalfont,labelsep=colon,strut=off} 
\floatstyle{ruled}
\newfloat{listing}{tb}{lst}{}
\floatname{listing}{Listing}
\title{Evaluate Geometry of Radiance Fields with Low-frequency Color Prior}
\newcommand{\correspondingauthor}{\thanks{Corresponding author.}}
\author{
    Qihang Fang\textsuperscript{\rm 1,2,}\equalcontrib, 
    Yafei Song\textsuperscript{\rm 3,}\equalcontrib, 
    Keqiang Li\textsuperscript{\rm 1,2}, 
    Li Shen\textsuperscript{\rm 3}, 
    Huaiyu Wu\textsuperscript{\rm 1}, 
    Gang Xiong\textsuperscript{\rm 1,}\correspondingauthor, 
    Liefeng Bo\textsuperscript{\rm 3}
}
\affiliations{
    \textsuperscript{\rm 1}State Key Laboratory of Multimodal Artificial Intelligence Systems, Institute of Automation,
Chinese Academy of Sciences\\
    \textsuperscript{\rm 2}School of Artificial Intelligence, University of Chinese Academy of Sciences\\
    \textsuperscript{\rm 3}Alibaba Group\\

    fangqihang2020@ia.ac.cn, huaizhang.syf@alibaba-inc.com, likeq98@gmail.com, lshen.lsh@gmail.com,\\ huaiyu.wu@ia.ac.cn, gang.xiong@ia.ac.cn, liefeng.bo@alibaba-inc.com
}

\usepackage{bibentry}

\begin{document}

\maketitle

\begin{abstract}
A radiance field is an effective representation of 3D scenes, which has been widely adopted in novel-view synthesis and 3D reconstruction.
It is still an open and challenging problem to evaluate the geometry, i.e., the density field, as the ground-truth is almost impossible to obtain.
One alternative indirect solution is to transform the density field into a point-cloud and compute its Chamfer Distance with the scanned ground-truth.
However, many widely-used datasets have no point-cloud ground-truth since the scanning process along with the equipment is expensive and complicated.
To this end, we propose a novel metric, named Inverse Mean Residual Color (IMRC), which can evaluate the geometry only with the observation images.
Our key insight is that the better the geometry, the lower-frequency the computed color field.
From this insight, given a reconstructed density field and observation images, we design a closed-form method to approximate the color field with low-frequency spherical harmonics, and compute the inverse mean residual color. 
Then the higher the IMRC, the better the geometry.
Qualitative and quantitative experimental results verify the effectiveness of our proposed IMRC metric.
We also benchmark several state-of-the-art methods using IMRC to promote future related research. Our code is available
at \url{https://github.com/qihangGH/IMRC}.
\end{abstract}

\section{Introduction}
\label{sec:intro}

\begin{figure}[t]
  \centering
   \includegraphics[width=1.\linewidth]{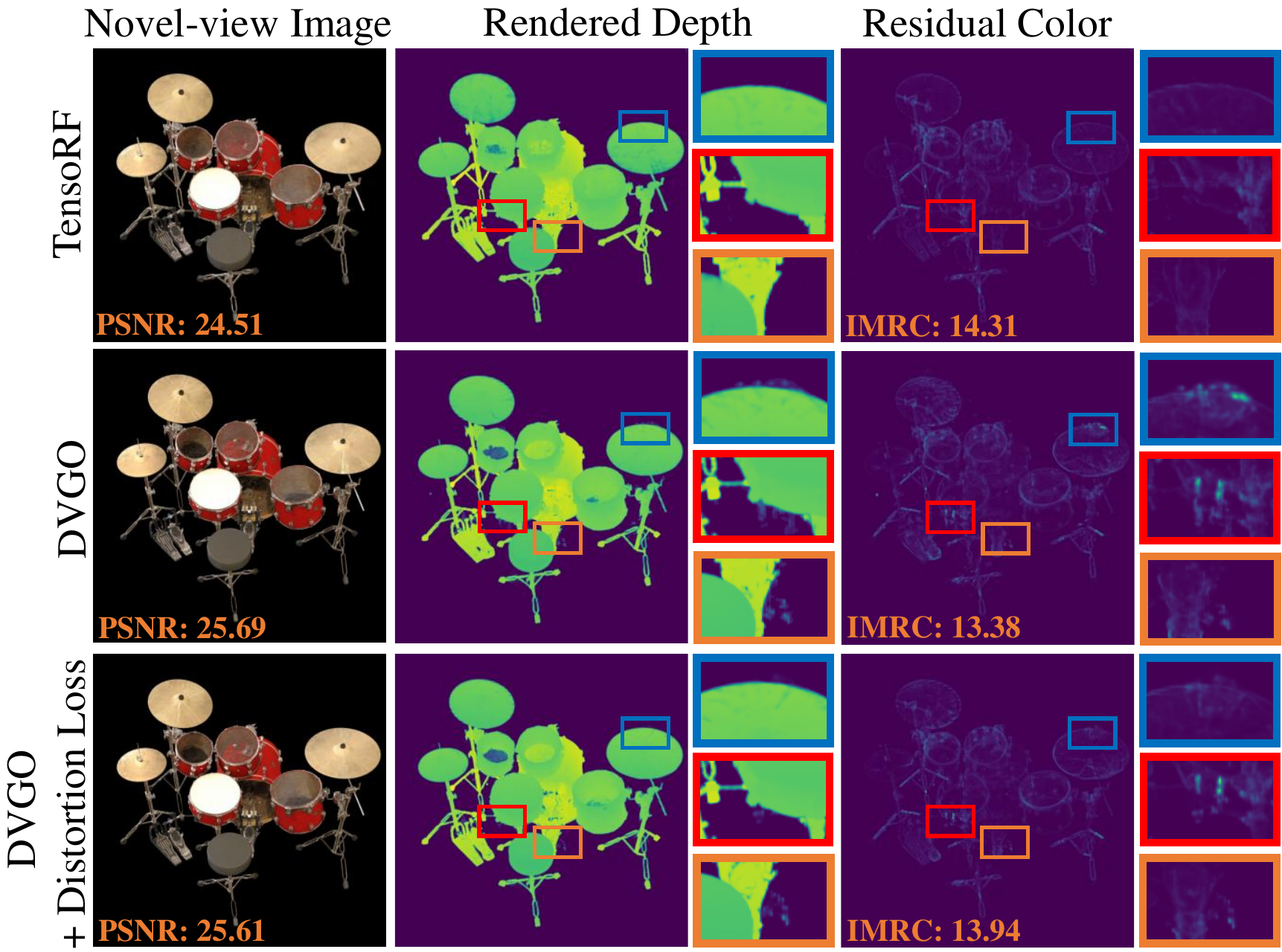}
\vspace{-2em}
   \caption{One example of novel-view images, rendered depth from reconstructed density fields, and residual color of TensoRF~\cite{AnpeiChen2022TensoRFTR}, DVGO~\cite{ChengSun2022DVGO}, and DVGO~\cite{ChengSun2022DVGO} + Distortion Loss~\cite{barron2022mipnerf360}. From top two rows, although DVGO achieves a better PSNR$\uparrow$, its geometry is qualitatively worse than TensoRF. From the bottom row, distortion loss could be qualitatively verified to improve the geometry. Our IMRC$\uparrow$ quantitatively evaluates these results correctly.}
   \label{fig:teaser}
\vspace{-2em}
\end{figure}

A radiance field has been popular for representing 3D objects or scenes since the Neural Radiance Field (NeRF) \cite{BenMildenhall2020NeRFRS} demonstrated that the performance of novel-view synthesis could be dramatically improved benefiting from it.
From this path-breaking work, many works aimed at improving NeRF from several aspects, such as various challenging scenarios \cite{KaiZhang2020NeRF++AA, RicardoMartinBrualla2020NeRFIT, peng2021animatable, barron2021MipNeRF, mildenhall2022NeRFdark}, inference speed \cite{AlexYu2021PlenOctreesFR, PeterHedman2021BakingNR, Reiser2021KiloNeRF, garbin2021fastnerf, chen2022mobilenerf}, training efficiency \cite{jaxnerf2020github, Hu2022EfficientNeRF, AlexYu2022PlenoxelsRF, ChengSun2022DVGO, sun2022DVGOv2, muller2022instantNGP, AnpeiChen2022TensoRFTR, wang2023mixed_iccv23, wang2023masked_nips23}, generalization ability \cite{yu2021pixelnerf, wang2021IBRNet, chen2021mvsnerf}, and so on.
Some other works proposed restoring the surfaces via reconstructing the latent radiance field by inverse volume rendering \cite{Wang2021NeuS, Long2022SpareseNeuS, Oechsle2021UNISURF, Sun2022SIGGRAPH, Zhang2021NeRS, Wang2022InstanNGP_NeuS, wu2022ObjectSDF, Wu2023OjbectSDF++}.

As the ground-truth of a radiance field is hard to obtain, we could not directly evaluate the reconstructed result.
Alternatively, novel-view synthesis methods usually measure the similarity between the rendered image and its observed counterpart using image quality metrics such as peak signal-to-noise ratio (PSNR).
As shown in the top two rows of \cref{fig:teaser}, this metric may be enough to evaluate the synthetic images but not appropriate to evaluate the geometry.

Besides novel-view synthesis methods, some algorithms aim at improving the reconstructed geometry, \textit{e.g.}, distortion loss~\cite{barron2022mipnerf360}. And a lot of works exploit radiance fields for 3D reconstruction~\cite{Wang2021NeuS, Long2022SpareseNeuS, Oechsle2021UNISURF, Sun2022SIGGRAPH, Zhang2021NeRS, Zhang2022IRON, wu2022voxurf, Wang2022InstanNGP_NeuS, wu2022ObjectSDF, Wu2023OjbectSDF++}.
These methods need a proper metric to quantitatively evaluate the geometry result.
To this end, surface reconstruction methods, \textit{e.g.}, \cite{Wang2021NeuS, Long2022SpareseNeuS}, usually transform the density field into a point-cloud using the marching cubes algorithm and compute Chamfer Distance (CD) with the scanned ground-truth.
However, the scanning process along with the equipment is expensive and complicated.
Therefore, few datasets have this ground-truth, and many widely-used ones do not.

To alleviate difficulties above, our key observation is that the color of any point in an ideal radiance field tends to be low-frequency if it is on the ground-truth surface of an object.
This phenomenon is named as low-frequency color prior.
Besides this, we adopt a closed-form method to compute the color field given the observation images and density field of a scene.
We name the result as the computed color field to distinguish it from the reconstructed one given by the radiance field reconstruction method.
Via exploring the computed color field, we find that the low-frequency color prior may be invalid for the points with inaccurate density values.
Therefore, the density field could be evaluated by the mean frequency of a computed color field.

However, it is difficult to directly evaluate the color frequency even for a single point, not to mention the whole field.
To this end, we approximate the color with low-frequency spherical harmonics and compute the residual color. Then, smaller mean residual color (MRC) implies lower color frequency, which indicates better geometry.
Moreover, as the MRC is usually very small, it is not convenient for presentation and comparison. 
Thus, we transform it into decibel (dB) as PSNR does and name it as inverse mean residual color (IMRC).
Then the higher the IMRC, the better the geometry.
As demonstrated in \cref{fig:teaser}, our IMRC metric could quantitatively evaluate the geometry correctly.

Our main contributions are concluded as follows:
1) We present the low-frequency color prior via analysing the ideal radiance field. We further find that this prior may be invalid for the computed color field if the density field is inaccurate.
2) To quantitatively evaluate the prior, we propose to approximate the color with low-frequency spherical harmonics and design the inverse mean residual color as a new metric. Qualitative and quantitative experimental results verify its effectiveness.
3) We further benchmark several state-of-the-art radiance field reconstruction methods using inverse mean residual color to promote future related research.

\section{Related Work}
\label{sec:related_work}

We firstly review two types of radiance field reconstruction works, and then discuss the geometry metrics.

\textbf{Novel View Synthesis}.
Mildenhall \textit{et al.} \cite{BenMildenhall2020NeRFRS} proposed NeRF to synthesize novel-view images from posed images and dramatically improved the performance.
They adopted a multilayer perceptron (MLP) to present the radiance field, which contains two components, \textit{i.e.}, the density field and color field.
Each point in the density field has a scalar controlling how much the color is accumulated.
Each point in the color field encodes the view-dependent color.
The image can be rendered using the volume rendering algorithm~\cite{max1995optical}.
From this path-breaking work, there have been many works aiming at improving NeRF from several aspects, such as various challenging scenarios \cite{KaiZhang2020NeRF++AA, RicardoMartinBrualla2020NeRFIT, barron2021MipNeRF, peng2021animatable, jain2021putting, mildenhall2022NeRFdark, Tancik2022BlockNeRF, Weng2022HumanNeRFMono, Zhao2022HumanNeRFSparse, Shao2022DoubleField, Niemeyer2022RegNeRF}, inference speed \cite{AlexYu2021PlenOctreesFR, PeterHedman2021BakingNR, Reiser2021KiloNeRF, garbin2021fastnerf, chen2022mobilenerf}, training efficiency \cite{jaxnerf2020github, Hu2022EfficientNeRF, AlexYu2022PlenoxelsRF, ChengSun2022DVGO, sun2022DVGOv2, muller2022instantNGP, AnpeiChen2022TensoRFTR, wang2023mixed_iccv23, wang2023masked_nips23}, generalization ability \cite{yu2021pixelnerf, wang2021IBRNet, chen2021mvsnerf}, and so on.

Since these works focused on novel view synthesis tasks, to evaluate the results, they usually adopted image quality metrics, such as PSNR, SSIM, and LPIPS.
These metrics are suitable for the task. 
However, it remains questionable how to evaluate the radiance field itself and whether these metrics are adequate. Evaluating the radiance field directly is challenging due to the difficulty in obtaining its ground-truth. Existing works have not well explored this problem.

\textbf{Surface Reconstruction}.
Some other works proposed to restore the surfaces via reconstructing the latent radiance field by inverse volume rendering \cite{Wang2021NeuS, Oechsle2021UNISURF, Sun2022SIGGRAPH, wu2022voxurf, Zhang2022IRON, wu2022ObjectSDF, Wu2023OjbectSDF++}.
These methods usually obtain the surfaces via transforming the latent density field into an occupancy field or signed distance field (SDF).
Then the Chamfer Distance between the points on the surfaces and the ground-truth can be computed to evaluate the results.
However, it is expensive and complicated to set up the hardware environment and scan the ground-truth point-cloud.
Moreover, during the transformation process, some information would be discarded as the transformed result only contains the iso-surface.
Therefore, the CD metric may not be suitable to evaluate a density field.
In contrast, IMRC is not suitable to evaluate the surfaces, but it could evaluate the density field. 

\textbf{Geometry Metrics}.
Due to the unavailable density field's ground-truth, we can not use direct evaluation metrics, such as mean squared error or mean absolute error.
Therefore, only indirect metrics could be considered.
In previous Structure-from-Motion (SfM) systems, \textit{e.g.}, \cite{Hartley2004MVG, Snavely2006PhotoTourism, Snavely2008Modeling, Agarwal2009BuildingRome}, the mean re-projection error is well-known and widely adopted as a metric and optimization objective to evaluate the reconstructed structures and motions.
The mean re-projection error is defined as the mean distance between each observed image feature point and the re-projection point of its reconstructed 3D point.
In other words, it could evaluate the reconstructed geometry without the corresponding ground-truth.
Inspired by this, we argue that the geometry of a radiance field also could be evaluated only with observation images.

\section{Low-frequency Color Prior}
\label{sec:prior}


According to NeRF~\cite{BenMildenhall2020NeRFRS}, a radiance field $\mathcal{F}$ is defined on a 3D space $\mathbf{V}$, which has two components, \textit{i.e.}, the density field $\mathcal{F}^{\sigma}$ and the color field $\mathcal{F}^{c}$.
For a 3D point $\mathbf{v}$ in the space $\mathbf{V}$, the density $\mathcal{F}^{\sigma}_{\mathbf{v}}$ is a scalar that controls how much color of this point is accumulated.
And the color $\mathcal{F}^{c}_{\mathbf{v}}$ encodes all view-dependent color information.
For a specific 3D direction $\mathbf{d}$, the color vector is denoted as $\mathcal{F}^{c}_{\mathbf{v}, \mathbf{d}}$.
To render the color observed from a ray $\mathbf{r}(t) = \mathbf{o} + t \mathbf{d}$ with near and far bounds $t^n, t^f$, we can follow the function
\begin{equation}
  \label{eq:rendering_int}
  {C}(\mathbf{r}) = \int_{t^n}^{t^f} T(t) \mathcal{F}^{\sigma}_{\mathbf{r}(t)} \mathcal{F}^{c}_{\mathbf{r}(t), \mathbf{d}} ~d t,
\end{equation}
where $\mathbf{o}$ is the camera original point, $\mathbf{d}$ is the direction from $\mathbf{o}$ to the corresponding pixel center, and the transmittance
\begin{equation}
  \label{eq:Trans_int}
  T(t) = \exp \left( \int_{t^n}^{t} -\mathcal{F}^{\sigma}_{\mathbf{r}(s)} ~d s \right).
\end{equation}
To numerically estimate this continuous integral, NeRF resorts to quadrature \cite{max1995optical}.
Then rendering equation is
\begin{equation}
  \label{eq:rendering_disc}
  \hat{{C}}(\mathbf{r}) = \sum_{i=1}^{N} \hat{T}(t_i) 
  \left(1 - \exp(-\mathcal{F}^{\sigma}_{\mathbf{r}(t_i)} \delta_{\textbf{r}(t_i)}) \right)
  \mathcal{F}^{c}_{\mathbf{r}(t_i), \mathbf{d}},
\end{equation}
where $N$ is the number of sample points along the ray,
\begin{equation}
  \label{eq:Trans_disc}
  \hat{T}(t_i) = \exp\left(- \sum_{j=1}^{i-1} \mathcal{F}^{\sigma}_{\mathbf{r}(t_j)} \delta_{\textbf{r}(t_j)} \right),
\end{equation}
and $\delta_{\textbf{r}(t_i)}$ is the distance between the point $\textbf{r}(t_i)$ and its previous neighbour $\textbf{r}(t_{i-1})$. Specifically, $\delta_{\textbf{r}(t_1)} = 0$.

\begin{figure}[t]
  \centering
   \includegraphics[width=1.0\linewidth]{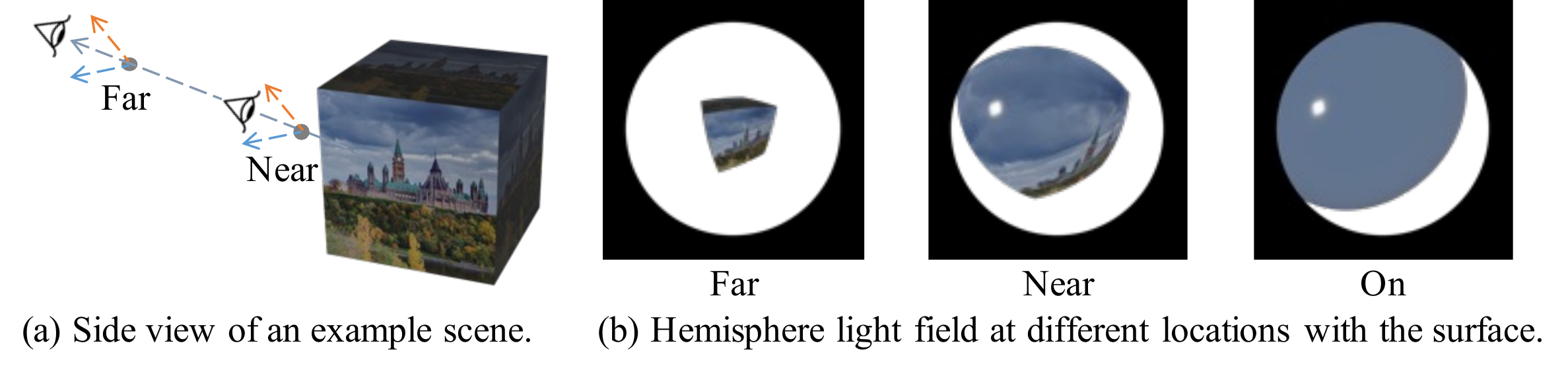}
\vspace{-2em}
   \caption{The color frequency tends to be lower and lower when the point approaches a surface.}
   \label{fig:color_prior}
\vspace{-1.5em}
\end{figure}

To illustrated the low-frequency color prior, we construct an example scene in \cref{fig:color_prior}~(a), which consists of a cube.
For better understanding, we also refer to the well-known related conception, \textit{i.e.}, light field, which describes the amount of light flowing in every direction through every point in space~\cite{ng2005fourier}. 
The difference between a light field and the color field of a radiance field is that, if a point has a zero density, its color in the color field could be arbitrary, but its color must be determined in the light field. 
If a point has a non-zero density, the colors in these two fields would be identical.
Besides, a light field has no density information.
As illustrated in \cref{fig:color_prior}~(b), the color frequency of a point tends to be lower when it approaches a surface. As the ground-truth density field should be the surface of the cube, and we do not care about the density values inside the surface as they do not affect the shape, a straight-forward hypothesis is that, for an ideal radiance field, the transmittance-density-weighted mean of color frequency should be small.
This prior widely exists for most types of surface whether it is rough or glossy.

\begin{figure}[t]
  \centering
   \includegraphics[width=1.0\linewidth]{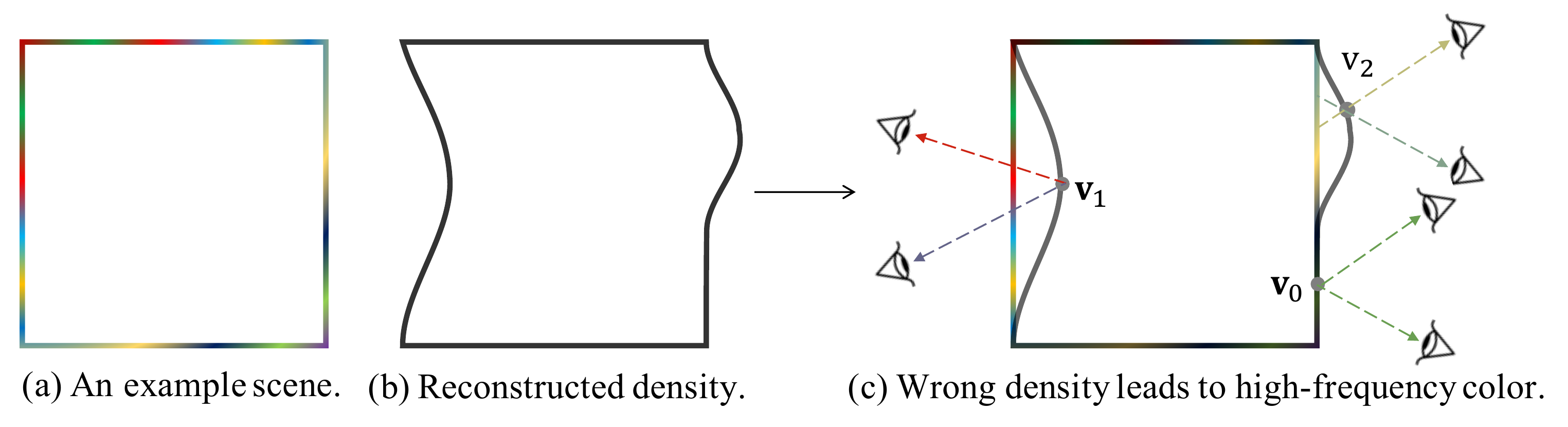}
\vspace{-1em}
   \caption{As the points $\textbf{v}_1$ and $\textbf{v}_2$ demonstrate, inaccurate surfaces lie inside or outside the ground-truth surface will lead to higher-frequency colors.}
   \label{fig:wrong_density}
\vspace{-1.5em}
\end{figure}


Moreover, we find that a wrong shape will break the low-frequency color prior.
To illustrate this, we present an example scene in \cref{fig:wrong_density}~(a), its reconstructed density field in \cref{fig:wrong_density}~(b), and combine the scene and the reconstructed density field in \cref{fig:wrong_density}~(c).
Intuitively, as demonstrated in \cref{fig:wrong_density}~(c), the observation color of a point from a ray should be identical with the color at the intersection of this ray and ground-truth surface.
The color frequency of the points on wrong surfaces that lie inside ($\textbf{v}_1$) or outside ($\textbf{v}_2$) the ground-truth tends to be higher than that of the ground-truth surface ($\textbf{v}_0$).
Note that we use a 2D graph here for simplicity. All analyses and conclusions are applicable for 3D cases.

From these observations, if we have the color frequency field computed from a density field and observation images, the transmittance-density-weighted mean color frequency
should be small for a good reconstructed density field.
However, it is not easy to figure out the color frequency of even a single point, not to mention the frequency field.
To alleviate this difficulty, we resort to the conjugate problem.
According to the frequency domain transformation theory, if a signal is low-frequency, we could well approximate the signal with a group of low-frequency basis, and the residual would be quite small.
Otherwise, the residual would be large.
Therefore, the residual could be adopted to replace the frequency. Suppose that the residual color of a point $\mathbf{v}$ at a direction $\mathbf{d}$ is $\mathcal{F}^{r}_{\mathbf{v}, \mathbf{d}}$, and the transmittance of $\mathbf{v}$ along $\mathbf{d}$ is $T_{\mathbf{v},\mathbf{d}}$,
the transmittance-density-weighted mean residual color
\begin{equation}
  \label{eq:residual_int}
  {r}_{\mathbf{V}} = \frac{\int_{\mathbf{v}} \int_{\mathbf{d}} T_{\mathbf{v},\mathbf{d}}\mathcal{F}^{\sigma}_{\mathbf{v}} \mathcal{F}^{r}_{\mathbf{v}, \mathbf{d}} ~d \mathbf{v} d\mathbf{d}}{\int_{\mathbf{v}} \int_{\mathbf{d}} T_{\mathbf{v},\mathbf{d}}\mathcal{F}^{\sigma}_{\mathbf{v}} ~d \mathbf{v} d\mathbf{d}} 
\end{equation}
should be small for a good reconstructed density field. We use the product of transmittance and density, i.e., $T_{\mathbf{v},\mathbf{d}}\mathcal{F}^{\sigma}_{\mathbf{v}}$, as the residual color's weight, which is the same as the color weight used in the rendering equation \eqref{eq:rendering_int}. In this way, only the points near the reconstructed surface are of the interest. The high color frequency of a point inside the reconstructed surface is not considered, as it does not affect the shape.

\section{Inverse Mean Residual Color}
\label{sec:method}

\begin{figure}[t]
  \centering
  \includegraphics[width=1.\linewidth]{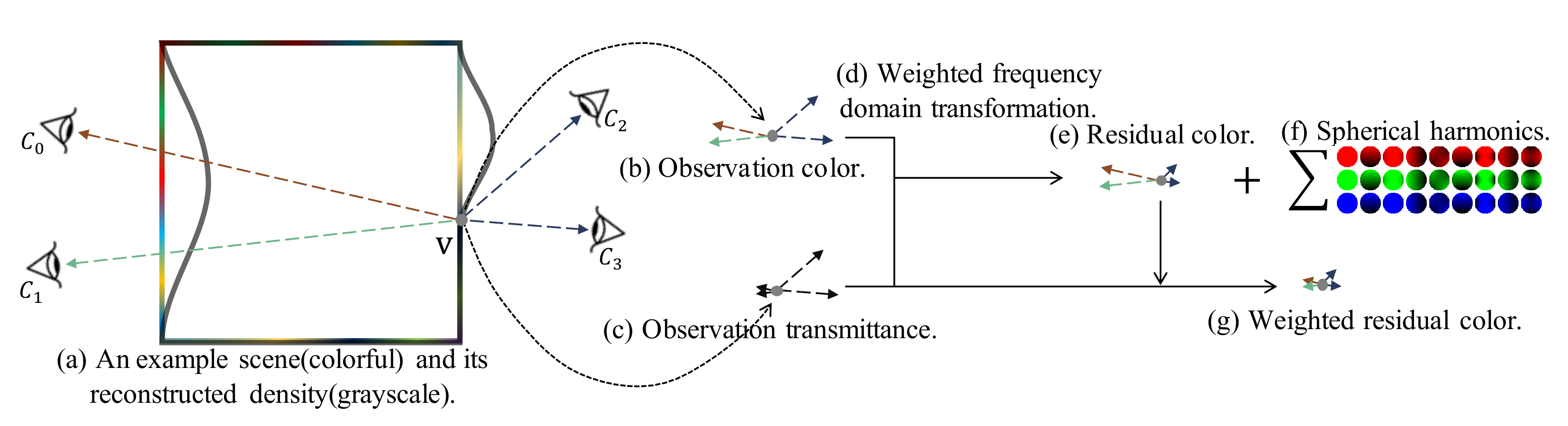}
\vspace{-2em}
  \caption{Residual color computation. (a) demonstrates an example scene and its reconstructed density field. For a point $\textbf{v}$, we can obtain its observation colors (b) according to the captured images. To tackle the occlusion, the transmittance between the point $\textbf{v}$ and each camera $C$ could be taken as the confidence of each observation, \textit{i.e.}, (c). Based on (b) and (c), the color could be weighted-transformed (d) into frequency domain (f) with the residual color (e). Due to the confidence (c), the residual color will be large if its corresponding confidence is low. To this end, the final residual color (g) should be weighted by observation confidence (c).}
  \label{fig:pipeline}
\vspace{-1em}
\end{figure}

In this section, we present the algorithm to compute the inverse mean residual color.
The difficulty is that we could not use the color field reconstructed by the radiance field reconstruction method.
If so, the geometry is not evaluated individually.
To overcome this, we propose to utilize the observation images.
For a 3D point, its projection points on all images are its observations.
We can approximate its observations with a group of low-frequency basis, and the residual color could be obtained.
We illustrate the whole pipeline in \cref{fig:pipeline} and first introduce the observation acquisition process, then transform them into the low-frequency domain, and compute the IMRC at last.

\subsection{Observation Acquisition}
\label{sec:observations}

As illustrated in \cref{fig:pipeline}~(a), for any 3D point $\mathbf{v}$ in the scene, given the reconstructed density field $\mathcal{F}^{\sigma}$ and $K$ cameras, we could obtain at most $K$ observations corresponding to the observed images.
For a specific camera, we denote its projection matrix and original point as $\mathbf{P}_k$ and $\mathbf{o}_{k}$ respectively.
Then the point $\mathbf{v}$ could be projected on the image plane following the projection equation
\begin{equation}
  \label{eq:projection}
  v = \mathbf{P}_k \mathbf{v},
\end{equation}
where $v$ is the coordinate on an image.
Its observed color $\mathbf{c}_k$ as illustrated in \cref{fig:pipeline}~(b) could be calculated using bi-linear interpolation. Besides the color, we should also record the observation direction since the view-dependent color is a function of it.
The observation direction is calculated as
\begin{equation}
  \label{eq:direction}
  \mathbf{d}_k = \frac{\mathbf{o}_k - \mathbf{v}}{\lVert \mathbf{o}_k - \mathbf{v} \rVert},
\end{equation}
where $\lVert \cdot \rVert$ is the $L_2$ normalization operator.

It is important to note that, due to the occlusion, observations from different cameras should not be regarded equally.
As illustrated in \cref{fig:pipeline}~(c), it is easy to understand that, if the accumulated density between the 3D point $\mathbf{v}$ and camera original point $\mathbf{o}_k$ is small, the corresponding observation confidence should be high, and vice versa.
Coincidentally, this measurement is actually the transmittance between the 3D point and the camera.
Therefore, we directly take the transmittance to measure the confidence of each observation, which could be calculated as
\begin{equation}
  \label{eq:confidence_int}
  T_{k} = \exp\left( \int_{0}^{\lVert \mathbf{o}_k - \mathbf{v} \rVert} -\mathcal{F}^{\sigma}_{\mathbf{r}(s \lvert \mathbf{v}, \mathbf{d}_k)} ~ds \right),
\end{equation}
where $\mathbf{r}(s \lvert \mathbf{v}, \mathbf{d}_k) = \mathbf{v} + s \mathbf{d}_k$.
However, this continuous integral is not easy to be computed for a computer.
Therefore, we numerically estimate it following NeRF \cite{BenMildenhall2020NeRFRS} as
\begin{equation}
  \label{eq:confidence_disc}
  \hat{T}_{k} = \exp\left( - \sum_{i=1}^{N} \mathcal{F}^{\sigma}_{\mathbf{r}(t_i \lvert \mathbf{v}, \mathbf{d}_k)} \delta_{\textbf{r}(t_i \lvert \mathbf{v}, \mathbf{d}_k)} \right),
\end{equation}
where $N$ is the number of sample points, and $\mathbf{r}(t_i \lvert \mathbf{v}, \mathbf{d}_k) = \mathbf{v} + t_i \mathbf{d}_k$ is the $i$-th point.

Overall, for any 3D point, we will have $K$ observations where each has three cells, \textit{i.e.}, color, direction, and confidence, which is denoted as $\langle \mathbf{c}_k, \mathbf{d}_k, \hat{T}_k \rangle$.
Specially, if the point is outside the viewing cone of the $k$-th camera, the confidence $\hat{T}_k$ should be set as $0$.

\subsection{Frequency Domain Transformation}
\label{sec:transformation}

In this subsection, we approximate the observations with a group of low-frequency basis.
A natural choice is to use spherical harmonics (SH), since they have been widely adopted in computer graphics to represent low-frequency colors \cite{Basri2003lambertianSH, Ramamoorthi2001relationshipSH} and also have been successfully applied to radiance field reconstruction \cite{AlexYu2021PlenOctreesFR, AlexYu2022PlenoxelsRF}.
More details about SH could be found in \cite{Green2003SH}. 
In the following, we briefly review the frequency coefficient estimation process following the well-known Monte Carlo method and present the difference under our situation.
Note that, partial contents of this subsection have been presented in our previous work \cite{fang2023reducing_nips23}.

From the frequency domain transformation theory, given a signal $F$ defined on the unit sphere $\mathbf{S}$, we can obtain its coefficient $h_\ell^m$ of the basis function $Y_\ell^m$ as
\begin{equation}
\begin{aligned}
  \label{eq:sh_int}
  h_\ell^m & = \int_{\mathbf{d} \in \mathbf{S}} \frac{F({\mathbf{d}})  Y_\ell^m(\mathbf{d})}{p(\mathbf{d})}  p(\mathbf{d}) ~ d {\mathbf{d}} \\
  & = \mathop{\mathbb{E}}\left(\frac{F({\mathbf{d}}) Y_\ell^m(\mathbf{d})}{p(\mathbf{d})}\right) = 4 \pi \mathop{\mathbb{E}}\left(F({\mathbf{d}}) Y_\ell^m(\mathbf{d})\right) \\
\end{aligned},
\end{equation}
where $\ell \in \mathbb{N}, m \in \mathbb{N} \cap \left[-\ell, \ell\right]$ is the degree and order of the SH basis, $p(\mathbf{d})$ is the sampling probability of direction $\mathbf{d}$, $\mathop{\mathbb{E}}(\cdot)$ is the expectation. As $\mathbf{d}$ is sampled evenly on the unit sphere $\mathbf{S}$, $p(\mathbf{d}) \equiv \frac{1}{4\pi}$.
Based on this, given $K$ observations which are obtained in the previous subsection, we can estimate the coefficient as
\begin{equation}
  \label{eq:sh_disc}
  \hat{h}_\ell^m = 4 \pi \frac{1}{K} \sum_{k=1}^K \mathbf{c}_k  Y_\ell^m(\mathbf{d}_k).
\end{equation}
However, this equation treats all observations equally, ignoring their varying confidence levels. To account for confidence $T_k$, the equation can be revised as
\begin{equation}
  \label{eq:sh_disc_weight}
  \hat{h}_\ell^m = 4 \pi \frac{1}{\sum_k T_k} \sum_{k=1}^K T_k  \mathbf{c}_k  Y_\ell^m(\mathbf{d}_k).
\end{equation}

Equation \eqref{eq:sh_disc_weight} above still has one problem. It implicitly assumes that the direction $\mathbf{d}$ distributes uniformly on the unit sphere $\mathbf{S}$. Therefore, the estimation process of each $\hat{h}_\ell^m$ is independent.
In computer graphics, this could be guaranteed via uniformly sampling the direction. 
However, this is not true under our settings as we cannot control the observation directions at all.
In practice, the observation directions usually are not uniformly distributed.

To overcome this, we should estimate the coefficients in turn and eliminate the influence of previous estimated coefficients.
Then Eq. \eqref{eq:sh_disc_weight} could be further updated as
\begin{equation}
  \label{eq:sh_disc_weight_history}
  \hat{h}_\ell^m = 4 \pi \frac{1}{\sum_k T_k} \sum_{k=1}^K T_k \cdot \tensor*[^m_\ell]{\tilde{\mathbf{c}}}{}_k  \cdot Y_\ell^m(\mathbf{d}_k),
\end{equation}
where
\begin{equation}
  \label{eq:residual_color_k}
  \tensor*[^m_\ell]{\tilde{\mathbf{c}}}{}_k = \mathbf{c}_k - \sum_{0 \leq i \leq \ell}^{-i \leq j < \tilde{m}} \hat{h}_i^j \cdot Y_i^j(\mathbf{d}_k),
\end{equation}
where $\tilde{m} = m$ if $i = \ell$, else $\tilde{m} = i + 1$.
In practice, the implementation involves incrementally increasing the degree and order while estimating the coefficients of the corresponding SH basis.
Before each estimation, we remove the color encoded by previous basis functions and only use the residual color $\tensor*[^m_\ell]{\tilde{\mathbf{c}}}{}_k$.


Note that, when the direction is uniformly distributed, \cref{eq:sh_disc_weight_history} is equivalent to \cref{eq:sh_disc} since the basis functions are orthogonal to each other.
Actually, \cref{eq:sh_disc_weight_history} is the general equation to transform a discrete signal from the original domain to another one. If with the uniform distribution and orthogonal basis, we have $T_k = 1, \sum_{k=1}^K Y_i^j(\mathbf{d}_k) \cdot Y_\ell^m(\mathbf{d}_k) = 0$, and \cref{eq:sh_disc_weight_history} could be transformed to \cref{eq:sh_disc} because
\begin{equation}
\begin{scriptsize}
\begin{aligned}
  \label{eq:uniform_weight_noweight}
  &\hat{h}_\ell^m = 4 \pi \frac{1}{K} \sum_{k=1}^K  \left(\mathbf{c}_k - \sum_{0 \leq i \leq \ell}^{-\ell \leq j < \tilde{m}} \hat{h}_i^j Y_i^j(\mathbf{d}_k)\right)  Y_\ell^m(\mathbf{d}_k) = \\
  &\frac{4 \pi}{K} \sum_{k=1}^K \mathbf{c}_k Y_\ell^m(\mathbf{d}_k) -
   \frac{4 \pi}{K} \sum_{k=1}^K \sum_{0 \leq i \leq \ell}^{-\ell \leq j < \tilde{m}} \hat{h}_i^j Y_i^j(\mathbf{d}_k) Y_\ell^m(\mathbf{d}_k)  
\end{aligned},
\end{scriptsize}
\end{equation}
and 
\begin{equation}
\begin{small}
\begin{aligned}
  \label{eq:uniform_zero}
  & \sum_{k=1}^K \sum_{0 \leq i \leq \ell}^{-\ell \leq j < \tilde{m}} \hat{h}_i^j Y_i^j(\mathbf{d}_k) Y_\ell^m(\mathbf{d}_k) = \\
  & \sum_{0 \leq i \leq \ell}^{-\ell \leq j < \tilde{m}} \hat{h}_i^j \sum_{k=1}^K  Y_i^j(\mathbf{d}_k) Y_\ell^m(\mathbf{d}_k) = 
  \sum_{0 \leq i \leq \ell}^{-\ell \leq j < \tilde{m}} \hat{h}_i^j \cdot 0 = 0
\end{aligned}.
\end{small}
\end{equation}
However, in our case, the directions are not uniformly distributed and weighted, therefore \cref{eq:sh_disc_weight_history} should be adopted.
The detailed proof and comparison results could be found in our previous work \cite{fang2023reducing_nips23}.

\subsection{Final Metric Computation}

With the calculated residual color at all locations, the MRC could be obtained following \cref{eq:residual_int}.
As it is a continuous integral, to numerically estimate it, we resort to quadrature \cite{max1995optical} once again.
Specifically, we discretize the 3D space $\mathbf{V}$ into a volume, and denote all voxel vertexes as $\mathcal{V}$.
Then the mean residual color could be estimated as
\begin{equation}
\begin{scriptsize}
\begin{aligned}
  \label{eq:residual_disc}
  \text{MRC} = \frac{  \sum_{\mathbf{v} \in \mathcal{V}} \sum_{k=1}^K T_{\mathbf{v},k} \left(1 - \exp\left(-\mathcal{F}^{\sigma}_{\mathbf{v}} \cdot \delta_{\mathbf{v}} \right) \right) \left( \tensor*[^{L}_{L}]{\tilde{\mathbf{c}}}{}_{\mathbf{v}, k} \right) ^ 2 }
  {\sum_{\mathbf{v} \in \mathcal{V}} \sum_{k=1}^K T_{\mathbf{v},k} \left(1 - \exp\left(-\mathcal{F}^{\sigma}_{\mathbf{v}} \cdot \delta_{\mathbf{v}} \right) \right)},
\end{aligned}
\end{scriptsize}
\end{equation}
where $\tensor*[^{L}_{L}]{\tilde{\mathbf{c}}}{}_{\mathbf{v}, k}$ is the final residual color at vertex $\mathbf{v}$ as defined in Eq. \eqref{eq:residual_color_k}, $T_{\mathbf{v}, k}$ is the transmittance along the ray that connects $\mathbf{v}$ and the $k$-th camera center $\mathbf{o}_k$ as defined in Eq. \eqref{eq:Trans_disc}, and $\delta_{\mathbf{v}}$ is the half voxel size.

Finally, we found that the MRC would be very small, which makes the value not convenient to be presented and compared.
To this end, we further transform it into decibel  as the PSNR metric does.
As the theoretically maximal residual color is $1$, the final metric could be obtained as
\begin{equation}
  \label{eq:mrc_log}
  \text{IMRC} = -10 \cdot \log_{10}\left(\text{MRC}\right).
\end{equation}
As the transformation includes an inverse operation, we denote it as IMRC shorted for Inverse Mean Residual Color. Then, the larger the IMRC, the better the geometry.

\section{Experiments}
\label{sec:exp}

In this section, we first verify the effectiveness of IMRC metric, then explore the influence of different settings.
Finally, we benchmark several state-of-the-art methods.

\subsection{Validation of IMRC}

To exploit the effectiveness of the proposed IMRC metric, we conduct a series of experiments on the DTU dataset \cite{Jensen2014DTU} as it provides the point-cloud ground-truth.
Though the point-cloud is essentially different from the radiance field, it still could be taken as a reference.
As previous surface reconstruction works did, \textit{e.g.}, \cite{Yariv2020IDR, Wang2021NeuS}, we use the selected $15$ scenes.
Each scene consists of $49$ or $64$ calibrated images and scanned point-cloud ground-truth.
We select $5$ or $6$ images from the total $49$ or $64$ images respectively as testing ones, and use the remaining as training ones.
The background of each image has been removed in advance.

For radiance field reconstruction methods, we select $4$ well-known methods, including JaxNeRF~\cite{BenMildenhall2020NeRFRS, jaxnerf2020github}, Plenoxels~\cite{AlexYu2022PlenoxelsRF}, DVGO~\cite{ChengSun2022DVGO}, and TensoRF~\cite{AnpeiChen2022TensoRFTR}, which have good performance.
We adopt the code released by the authors to optimize each scene.
Additionally, we add a transmittance loss to the JaxNeRF to supervise the transmittance of a ray according to the foreground mask.
After the optimization process, we only export the density field obtained by each method. 
Notably, JaxNeRF~\cite{BenMildenhall2020NeRFRS, jaxnerf2020github} models a continuous density field while other methods model discrete ones.
For a fair comparison, we discrete the density field of each method via sampling a $512^3$ density volume.
Then the IMRC metric could be calculated. The experimental results are presented in \cref{tab:dtu}, where the PSNR on testing images and the CD calculated with the point-cloud ground-truth are also reported. Specifically, we search the optimal threshold needed by the marching cubes algorithm for each method on each scene and report the lowest CD. Because the ground-truth of the density field is not available, we perform a user study inspired by previous work \cite{li2015DataDriven_iccv15}. Specifically, $3$ experts evaluate and rank the scene geometry produced by the $4$ methods from $1$ (best) to $4$ (worst) based on the depth images, residual colors, and meshes. They are unknown about the metric values and the method that produces the results. We report the mean rank results in \cref{tab:dtu}.

\begin{table}[t]
\tiny
\begin{center}
  \caption{The PSNR$\uparrow$/CD$\downarrow$/IMRC$\uparrow$/UserRank$\downarrow$ results of $4$ methods on the DTU dataset. (Rank \textcolor{red}{1st}, \textcolor[rgb]{0,0.5,0}{2nd}, \textcolor{blue}{3rd}, 4th)}%
  \label{tab:dtu}
\vspace{-1.5em}
  \begin{tabular}{@{\hspace{0mm}} l @{\hspace{0.2mm}} c @{\hspace{0.2mm}} c @{\hspace{0.2mm}} c @{\hspace{0.2mm}} c @{\hspace{0mm}}}

  \hline

  Method  & JaxNeRF  &  Plenoxels  & DVGO & TensoRF \\
  \hline
Scan24 & \textcolor{red}{29.19}/\textcolor{red}{1.415}/\textcolor{red}{20.11}/\textcolor{red}{1~~~} & \textcolor{black}{27.72}/\textcolor[rgb]{0,0.5,0}{1.592}/\textcolor{blue}{14.96}/\textcolor{blue}{3~~~} & \textcolor{blue}{27.96}/\textcolor{blue}{2.028}/\textcolor[rgb]{0,0.5,0}{17.45}/\textcolor[rgb]{0,0.5,0}{2~~~} & \textcolor[rgb]{0,0.5,0}{28.76}/\textcolor{black}{2.144}/\textcolor{black}{11.73}/\textcolor{black}{4} \\
Scan37 & \textcolor{red}{26.77}/\textcolor[rgb]{0,0.5,0}{1.502}/\textcolor{red}{15.43}/\textcolor{red}{1~~~} & \textcolor{blue}{26.22}/\textcolor{black}{1.813}/\textcolor{blue}{12.99}/\textcolor{blue}{3~~~} & \textcolor[rgb]{0,0.5,0}{26.47}/\textcolor{red}{1.466}/\textcolor[rgb]{0,0.5,0}{14.05}/\textcolor[rgb]{0,0.5,0}{2~~~} & \textcolor{black}{26.00}/\textcolor{blue}{1.778}/\textcolor{black}{~~8.86}/\textcolor{black}{4}\\
Scan40 & \textcolor{red}{29.43}/\textcolor{red}{1.495}/\textcolor{red}{20.74}/\textcolor{red}{1~~~} & \textcolor{blue}{28.72}/\textcolor{blue}{2.030}/\textcolor{blue}{15.75}/\textcolor{blue}{3~~~} & \textcolor{black}{28.38}/\textcolor[rgb]{0,0.5,0}{2.017}/\textcolor[rgb]{0,0.5,0}{17.43}/\textcolor[rgb]{0,0.5,0}{2~~~} & \textcolor[rgb]{0,0.5,0}{28.85}/\textcolor{black}{2.381}/\textcolor{black}{13.39}/\textcolor{black}{4}\\
Scan55 & \textcolor[rgb]{0,0.5,0}{31.06}/\textcolor{red}{0.637}/\textcolor{red}{19.55}/\textcolor{red}{1~~~} & \textcolor{blue}{30.50}/\textcolor[rgb]{0,0.5,0}{0.892}/\textcolor{blue}{15.55}/\textcolor{blue}{3~~~} & \textcolor{red}{31.66}/\textcolor{blue}{1.247}/\textcolor[rgb]{0,0.5,0}{19.37}/\textcolor[rgb]{0,0.5,0}{2~~~} & \textcolor{black}{29.57}/\textcolor{black}{1.765}/\textcolor{black}{~~9.78}/\textcolor{black}{4}\\
Scan63 & \textcolor{blue}{34.43}/\textcolor[rgb]{0,0.5,0}{1.689}/\textcolor{red}{20.68}/\textcolor{red}{1~~~} & \textcolor{black}{34.09}/\textcolor{blue}{1.931}/\textcolor{blue}{17.63}/\textcolor{blue}{3~~~} & \textcolor{red}{35.42}/\textcolor{red}{1.574}/\textcolor[rgb]{0,0.5,0}{19.94}/\textcolor[rgb]{0,0.5,0}{2~~~} & \textcolor[rgb]{0,0.5,0}{34.69}/\textcolor{black}{2.216}/\textcolor{black}{10.92}/\textcolor{black}{4}\\
Scan65 & \textcolor{black}{29.90}/\textcolor{red}{1.212}/\textcolor{red}{17.59}/\textcolor{red}{1.3} & \textcolor{red}{31.16}/\textcolor{blue}{1.548}/\textcolor{blue}{14.55}/\textcolor{blue}{3~~~} & \textcolor{blue}{30.63}/\textcolor[rgb]{0,0.5,0}{1.482}/\textcolor[rgb]{0,0.5,0}{17.32}/\textcolor[rgb]{0,0.5,0}{1.7} & \textcolor[rgb]{0,0.5,0}{30.63}/\textcolor{black}{1.947}/\textcolor{black}{10.78}/\textcolor{black}{4}\\
Scan69 & \textcolor{black}{29.25}/\textcolor{red}{1.306}/\textcolor{red}{19.07}/\textcolor{red}{1~~~} & \textcolor{red}{30.15}/\textcolor{black}{2.346}/\textcolor{blue}{15.40}/\textcolor{blue}{3~~~} & \textcolor[rgb]{0,0.5,0}{29.67}/\textcolor[rgb]{0,0.5,0}{1.489}/\textcolor[rgb]{0,0.5,0}{17.81}/\textcolor[rgb]{0,0.5,0}{2~~~} & \textcolor{blue}{29.43}/\textcolor{blue}{2.254}/\textcolor{black}{~~7.91}/\textcolor{black}{4}\\
Scan83 & \textcolor{black}{35.28}/\textcolor{red}{1.478}/\textcolor{blue}{15.98}/\textcolor{red}{1~~~} & \textcolor{red}{37.00}/\textcolor{blue}{2.245}/\textcolor[rgb]{0,0.5,0}{16.63}/\textcolor{blue}{2.7} & \textcolor[rgb]{0,0.5,0}{36.14}/\textcolor[rgb]{0,0.5,0}{1.610}/\textcolor{red}{17.95}/\textcolor[rgb]{0,0.5,0}{2.3} & \textcolor{blue}{35.96}/\textcolor{black}{2.712}/\textcolor{black}{~~7.95}/\textcolor{black}{4}\\
Scan97 & \textcolor{black}{28.02}/\textcolor{red}{1.600}/\textcolor{red}{17.10}/\textcolor{red}{1~~~} & \textcolor{red}{29.56}/\textcolor{black}{2.809}/\textcolor{blue}{15.29}/\textcolor{blue}{3~~~} & \textcolor{blue}{29.11}/\textcolor[rgb]{0,0.5,0}{1.629}/\textcolor[rgb]{0,0.5,0}{15.85}/\textcolor[rgb]{0,0.5,0}{2~~~} & \textcolor[rgb]{0,0.5,0}{29.25}/\textcolor{blue}{2.278}/\textcolor{black}{~~9.06}/\textcolor{black}{4}\\
Scan105 & \textcolor{black}{31.86}/\textcolor{red}{1.136}/\textcolor{red}{19.99}/\textcolor{red}{1~~~} & \textcolor{red}{33.36}/\textcolor{blue}{1.907}/\textcolor{blue}{16.67}/\textcolor{blue}{3~~~} & \textcolor[rgb]{0,0.5,0}{33.07}/\textcolor[rgb]{0,0.5,0}{1.352}/\textcolor[rgb]{0,0.5,0}{18.85}/\textcolor[rgb]{0,0.5,0}{2~~~} & \textcolor{blue}{32.92}/\textcolor{black}{2.219}/\textcolor{black}{~~9.19}/\textcolor{black}{4}\\
Scan106 & \textcolor[rgb]{0,0.5,0}{33.97}/\textcolor{red}{0.903}/\textcolor{red}{19.70}/\textcolor{red}{1~~~} & \textcolor{blue}{33.82}/\textcolor{blue}{2.317}/\textcolor{blue}{16.37}/\textcolor{blue}{3~~~} & \textcolor{red}{34.70}/\textcolor[rgb]{0,0.5,0}{1.595}/\textcolor[rgb]{0,0.5,0}{19.29}/\textcolor[rgb]{0,0.5,0}{2~~~} & \textcolor{black}{33.61}/\textcolor{black}{2.524}/\textcolor{black}{~~9.01}/\textcolor{black}{4}\\
Scan110 & \textcolor{blue}{32.96}/\textcolor[rgb]{0,0.5,0}{1.842}/\textcolor[rgb]{0,0.5,0}{16.00}/\textcolor[rgb]{0,0.5,0}{1.7} & \textcolor{black}{32.49}/\textcolor{blue}{3.349}/\textcolor{blue}{14.86}/\textcolor{blue}{3~~~} & \textcolor{red}{34.00}/\textcolor{red}{1.770}/\textcolor{red}{16.13}/\textcolor{red}{1.3} & \textcolor[rgb]{0,0.5,0}{33.67}/\textcolor{black}{3.396}/\textcolor{black}{~~7.99}/\textcolor{black}{4}\\
Scan114 & \textcolor{black}{29.44}/\textcolor{red}{1.091}/\textcolor{red}{18.89}/\textcolor{red}{1~~~} & \textcolor{red}{30.52}/\textcolor{blue}{1.657}/\textcolor{blue}{16.53}/\textcolor{blue}{3~~~} & \textcolor[rgb]{0,0.5,0}{30.47}/\textcolor[rgb]{0,0.5,0}{1.481}/\textcolor[rgb]{0,0.5,0}{18.18}/\textcolor[rgb]{0,0.5,0}{2~~~} & \textcolor{blue}{29.79}/\textcolor{black}{2.024}/\textcolor{black}{~~9.54}/\textcolor{black}{4}\\
Scan118 & \textcolor[rgb]{0,0.5,0}{36.68}/\textcolor{red}{0.998}/\textcolor[rgb]{0,0.5,0}{21.77}/\textcolor[rgb]{0,0.5,0}{1.7} & \textcolor{blue}{36.10}/\textcolor{black}{2.719}/\textcolor{blue}{17.82}/\textcolor{blue}{3~~~} & \textcolor{red}{37.58}/\textcolor[rgb]{0,0.5,0}{1.297}/\textcolor{red}{22.57}/\textcolor{red}{1.3} & \textcolor{black}{35.76}/\textcolor{blue}{2.436}/\textcolor{black}{11.34}/\textcolor{black}{4}\\
Scan122 & \textcolor{black}{35.06}/\textcolor{red}{0.761}/\textcolor[rgb]{0,0.5,0}{18.00}/\textcolor[rgb]{0,0.5,0}{2~~~} & \textcolor[rgb]{0,0.5,0}{36.42}/\textcolor{black}{2.349}/\textcolor{blue}{17.15} /\textcolor{blue}{3~~~} & \textcolor{red}{37.05}/\textcolor[rgb]{0,0.5,0}{1.244}/\textcolor{red}{20.62}/\textcolor{red}{1~~~} & \textcolor{blue}{36.18}/\textcolor{blue}{2.258}/\textcolor{black}{~~9.60}/\textcolor{black}{4}\\

\hline

Average & \textcolor{black}{31.55}/\textcolor{red}{1.271}/\textcolor{red}{18.71}/\textcolor{red}{1.18} & \textcolor[rgb]{0,0.5,0}{31.86}/\textcolor{blue}{2.100}/\textcolor{blue}{15.88}/\textcolor{blue}{2.98}& \textcolor{red}{32.15}/\textcolor[rgb]{0,0.5,0}{1.552}/\textcolor[rgb]{0,0.5,0}{18.19}/\textcolor[rgb]{0,0.5,0}{1.84}& \textcolor{blue}{31.67}/\textcolor{black}{2.289}/\textcolor{black}{~~9.80}/\textcolor{black}{4}\\
  \hline
  \end{tabular}
\end{center}
\vspace{-2.5em}
\end{table}





\begin{figure}[t]
   \includegraphics[width=1.\linewidth]{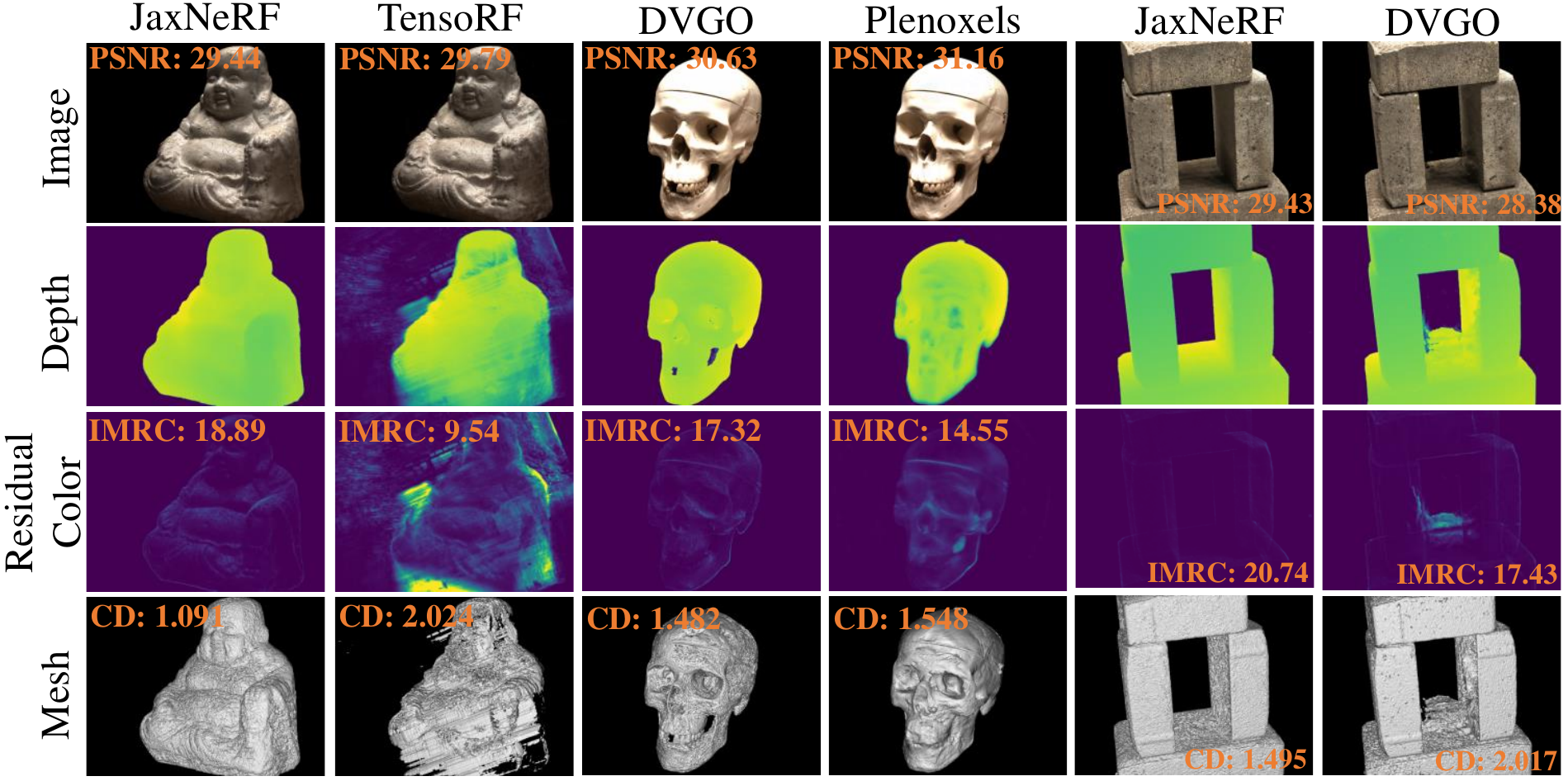}
\vspace{-2em}
   \caption{Examples of the CD$\downarrow$/IMRC$\uparrow$/UserRank$\downarrow$ consistent results on the DTU dataset. For each example, left is better. From left to right, Scan 114, 65, and 40.}
   \label{fig:PSNR_MRC_conflict}
\vspace{-1em}
\end{figure}

\begin{figure}[t]
   \includegraphics[width=1.\linewidth]{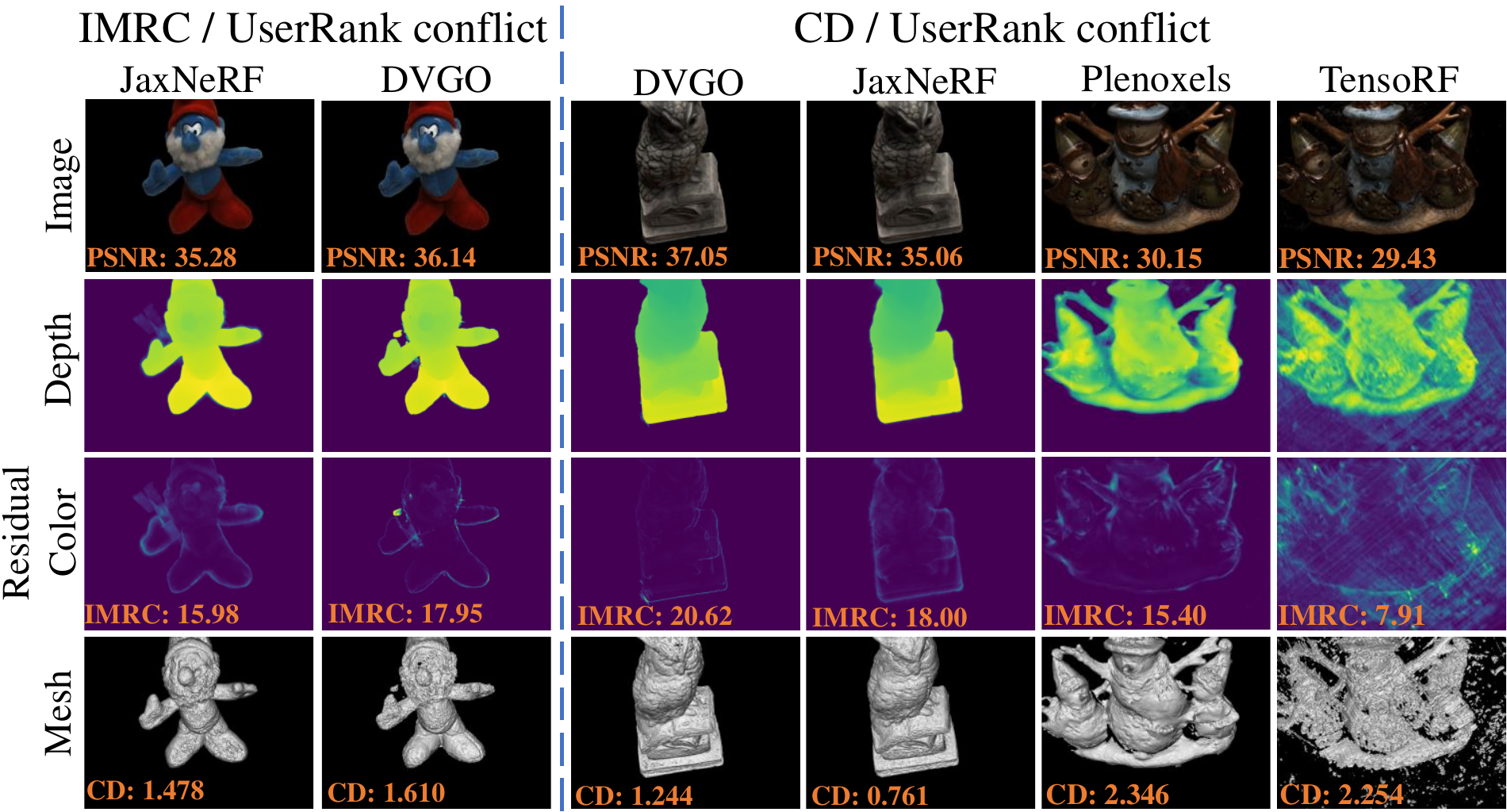}
\vspace{-1em}
   \caption{The IMRC$\uparrow$/UserRank$\downarrow$ and CD$\downarrow$/UserRank$\downarrow$ conflict results. For each example, left has better UserRank. From left to right, Scan 83, 122, and 69.}
   \label{fig:CD_MRC_conflicts}
\vspace{-1em}
\end{figure}

In most cases, IMRC, CD, and UserRank agree with each other. The average results (last row) in \cref{tab:dtu} show that they rank $4$ methods consistently. We showcase three CD/IMRC/UserRank consistent cases in \cref{fig:PSNR_MRC_conflict}. They also indicate that PSNR cannot well evaluate a density field.
Besides, there are 2 IMRC/UserRank conflict pairs and 11 CD/UserRank conflict pairs out of all $90 = 15 \cdot \tbinom{4}{2}$ pairs, respectively. In the left panel of \cref{fig:CD_MRC_conflicts}, one IMRC/UserRank conflict case is shown. The density field of the JaxNeRF is not sharp, and its surface is surrounded by many low density floaters. In contrast, the DVGO has a distinct floater as highlighted by the residual color. Although the IMRC of DVGO is better, such a distinct floater may lead to worse UserRank. The CD/UserRank conflicts mainly stem from two reasons. First, 
a NeRF model does not guarantee that its surface points all have the same density value. Therefore, by extracting one iso-surface at a typical density level with the marching cubes algorithm, some surface information and near surface floaters that have different density values are inevitably discarded. As a result, the CD 
for such a surface may not well reflect real geometry. As shown by the middle two columns of \cref{fig:CD_MRC_conflicts}. The JaxNeRF produces many low density floaters around the surface, as highlighted by its residual color, while these floaters are missing after applying the marching cubes algorithm. On the other hand, the iso-surface of DVGO is inferior because some of its surface points that have different density values are discarded, which leads to poor CD. Another reason is that an object mask is applied in CD calculation, so points that lie outside the mask are neglected. As shown in the last two columns of \cref{fig:CD_MRC_conflicts}, the TensoRF's mesh is qualitatively worse than Plenoxels', but lots of floaters in it are out of the mask. As a result, the CD of TensoRF is erroneously better than that of Plenoxels. The IMRC metric successfully evaluate the last two scenes and is consistent with UserRank. We visualize all the rest conflict cases in the Appendix. They can be analyzed similarly.


Overall, the qualitative and quantitative results verify the effectiveness of IMRC to evaluate a density field. On one hand, PSNR could only evaluate the radiance field as a whole. On the other hand, the calculation of CD suffers from a loss of information that only an iso-surface at one specific density level is considered. The IMRC metric, in contrast, is more suitable to evaluate the whole field, rather than only a surface, as it can deal with density values at all locations. It is also more consistent with UserRank. Moreover, the CD metric needs the point-cloud ground-truth, which also limits its application as this data is complicated and expensive to obtain and usually is not available.

\subsection{Different Settings of IMRC}

In this subsection, we exploit the influences of the parameters defined during IMRC computation.
There are only $2$ parameters that are needed to be configured manually, \textit{i.e.}, the SH degree and the volume resolution.

\begin{table}[b]
\tiny
\begin{center}
  \caption{IMRC results of the selected methods on the DTU dataset with different SH degrees, which indicates that the higher SH degree leads to better IMRC results.}%
  \label{tab:SH_degrees}
\vspace{-1em}
  \begin{tabular}{@{\hspace{2mm}} l @{\hspace{4mm}} c @{\hspace{4mm}} c @{\hspace{4mm}} c @{\hspace{4mm}} c @{\hspace{2mm}}}
  \hline
  SH Degree   & $0$ & $1$  & $2$ & $3$   \\
  SH Basis \# & $1$ & $4$  & $9$ & $16$  \\
  \hline
  JaxNeRF & 15.93 & 17.18 & 18.71 & 19.57 \\
  Plenoxels & 13.81 & 14.78 & 15.88 & 16.52
   \\
  DVGO           & 15.66 & 16.80 & 18.19 & 18.97
  \\
  TensoRF  
  & ~~9.00 & ~~9.40 & ~~9.80 & 10.06
 \\
  \hline
  Average   & 13.60 & 14.54 & 15.64 & 16.28 \\
  \hline
  \end{tabular}
\end{center}
\vspace{-1em}
\end{table}

The first and foremost parameter is the degree of SH used to compute the residual color for each point.
In \cref{tab:SH_degrees}, we present the IMRC results of the selected $4$ methods on the DTU dataset with different SH degrees.
We can see that a higher SH degree leads to a better IMRC result.
This is reasonable since the observations could be approximated better with more SH basis functions.
However, we could not set a very large SH degree, because, if so, the residual color will be very small at all positions in the field, which makes the metric indistinct to evaluate the geometry.
We also observe that the IMRC increases quickly when the SH degree is less than $2$.
And the increased margin becomes smaller starting from degree $3$.
Therefore, we set the SH degree as $2$ in all other experiments.
This setting is also usually used in previous works such as \cite{AlexYu2022PlenoxelsRF, AnpeiChen2022TensoRFTR}. Notably, the relative ranks of all methods remain unchanged with different SH degrees. This indicates that the IMRC is somehow stable with different SH degrees.

Notably, when the number of observations is small, we should decrease the SH degree.
This will happen for the methods aiming at reconstructing the radiance field from sparse-views, such as \cite{Niemeyer2022RegNeRF}.
According to the frequency domain transformation theory, the SH degree should be smaller than the number of observations.


The second parameter is the density volume's resolution.
As presented in \cref{tab:resolutions}, with the increasing resolution, the IMRC metric is slightly better.
The increase may come from the more accurate density field with higher space resolution.
We also observe that the relative ranks of all methods remain unchanged at different resolutions.
This indicates that the IMRC is also somehow stable with different resolutions.
As the average IMRC does not change from resolution $256^3$ to $512^3$, we adopt $512^3$ in other experiments.

\begin{table}[h]
\tiny
\vspace{-1em}
\begin{center}
  \caption{IMRC results of the selected methods on the DTU dataset with different density volume resolutions.}%
  \label{tab:resolutions}
\vspace{-1.5em}
  \begin{tabular}{@{\hspace{1mm}} l @{\hspace{3mm}} c @{\hspace{3mm}} c @{\hspace{3mm}} c @{\hspace{3mm}} c @{\hspace{3mm}} c @{\hspace{1mm}}}
  \hline
  Resolution  & $64^3$ & $128^3$  & $256^3$ & $512^3$ & $768^3$   \\
  \hline
  JaxNeRF 
  & 17.92 & 18.42 & 18.61 & 18.71 & 18.54 \\
  Plenoxels   & 15.96 & 16.04 & 16.04 & 15.88 & 16.00
  \\
  DVGO             & 17.05 & 17.72 & 18.14 & 18.19 & 18.27
  \\
  TensoRF    & ~~9.45 & ~~9.62 & ~~9.75 & ~~9.80 & ~~9.88
 \\
  \hline
  Average                                 & 15.09 & 15.45 & 15.64 & 15.64 & 15.67
\\
  \hline
  \end{tabular}
\end{center}
\vspace{-3em}
\end{table}

\subsection{Benchmarking State-of-the-arts}

\begin{figure}[h]
  \centering
   \includegraphics[width=1.0\linewidth]{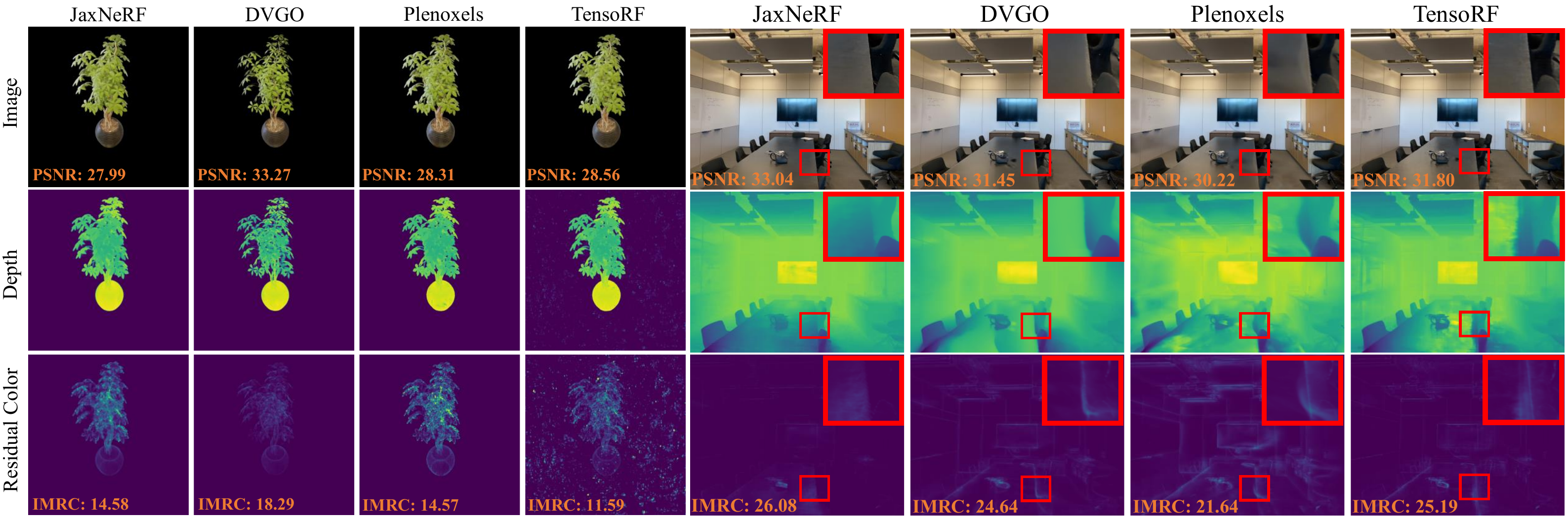}
\vspace{-2.0em}
   \caption{Example results on Ficus from the NeRF synthetic dataset and Room from the LLFF dataset.}
   \label{fig:results_synthetic_llff}
\vspace{-0.0em}
\end{figure}

Besides the DTU dataset, we also benchmark the $4$ state-of-the-art methods on NeRF Synthetic~\cite{BenMildenhall2020NeRFRS} and LLFF~\cite{Mildenhall2019LLFF} datasets.
For the NeRF Synthetic dataset, we use the released code and train on the black background as done in the DTU dataset.
For the LLFF dataset, we directly use the models released by the authors.
As there is no point-cloud ground-truth on NeRF Synthetic and LLFF datasets, we could not compute the CD metric and only present the PSNR and IMRC results in \cref{tab:synthetic} and \cref{tab:llff}, respectively.
We also visualize some results in \cref{fig:results_synthetic_llff}. 


\begin{table}[h]
\tiny
\begin{center}
\vspace{-0em}
  \caption{The PSNR$\uparrow$/IMRC$\uparrow$ results on NeRF synthetic dataset. (Rank \textcolor{red}{1st}, \textcolor[rgb]{0,0.5,0}{2nd}, \textcolor{blue}{3rd}, 4th)}%
  \label{tab:synthetic}
\vspace{-1.5em}
  \begin{tabular}{@{\hspace{0mm}} l @{\hspace{1mm}} c @{\hspace{2mm}} c @{\hspace{2mm}} c @{\hspace{2mm}} c @{\hspace{0mm}}}
  \hline
  Method  & JaxNeRF  & Plenoxels  & DVGO & TensoRF \\
  \hline
Chair & \textcolor{black}{30.28} / \textcolor[rgb]{0,0.5,0}{19.21} & \textcolor[rgb]{0,0.5,0}{30.66} / \textcolor{black}{17.60} & \textcolor{red}{33.79} / \textcolor{red}{21.59} & \textcolor{blue}{30.46} / \textcolor{blue}{19.10}\\
Drums & \textcolor{black}{24.16} / \textcolor[rgb]{0,0.5,0}{13.41} & \textcolor{blue}{24.26} / \textcolor{black}{12.65} & \textcolor{red}{25.69} / \textcolor{blue}{13.38} & \textcolor[rgb]{0,0.5,0}{24.51} / \textcolor{red}{14.32}\\
Ficus & \textcolor{black}{27.99} / \textcolor[rgb]{0,0.5,0}{14.58} & \textcolor{blue}{28.31} / \textcolor{blue}{14.57} & \textcolor{red}{33.27} / \textcolor{red}{18.29} & \textcolor[rgb]{0,0.5,0}{28.56} / \textcolor{black}{11.59}\\
Hotdog & \textcolor[rgb]{0,0.5,0}{35.49} / \textcolor{black}{~~4.59} & \textcolor{black}{35.11} / \textcolor{blue}{20.02} & \textcolor{red}{36.90} / \textcolor[rgb]{0,0.5,0}{21.16} & \textcolor{blue}{35.43} / \textcolor{red}{22.41}\\
Lego & \textcolor{blue}{31.41} / \textcolor{black}{~~3.73} & \textcolor{black}{30.87} / \textcolor{blue}{17.54} & \textcolor{red}{34.83} / \textcolor{red}{19.27} & \textcolor[rgb]{0,0.5,0}{31.83} / \textcolor[rgb]{0,0.5,0}{19.21}\\
Materials & \textcolor[rgb]{0,0.5,0}{29.67} / \textcolor[rgb]{0,0.5,0}{16.09} & \textcolor{black}{28.39} / \textcolor{blue}{15.15} & \textcolor{red}{29.94} / \textcolor{red}{17.16} & \textcolor{blue}{28.91} / \textcolor{black}{12.63}\\
Mic & \textcolor{blue}{31.22} / \textcolor{black}{~~2.15} & \textcolor[rgb]{0,0.5,0}{32.50} / \textcolor{red}{16.47} & \textcolor{black}{28.41} / \textcolor{blue}{~~8.09} & \textcolor{red}{32.88} / \textcolor[rgb]{0,0.5,0}{14.41}\\
Ship & \textcolor[rgb]{0,0.5,0}{28.76} / \textcolor{blue}{18.77} & \textcolor{black}{28.52} / \textcolor{black}{18.37} & \textcolor{red}{29.82} / \textcolor[rgb]{0,0.5,0}{19.08} & \textcolor{blue}{28.68} / \textcolor{red}{19.24}\\
\hline
Average & \textcolor{blue}{29.87} / \textcolor{black}{11.57} & \textcolor{black}{29.83} / \textcolor{blue}{16.55} & \textcolor{red}{31.58} / \textcolor{red}{17.25} & \textcolor[rgb]{0,0.5,0}{30.16} / \textcolor[rgb]{0,0.5,0}{16.61}\\
\hline
  \end{tabular}
\end{center}
\vspace{-1em}
\end{table}

\begin{table}[h]
\tiny
\begin{center}
\vspace{-1em}
  \caption{The PSNR$\uparrow$/IMRC$\uparrow$ results on LLFF dataset. (Rank \textcolor{red}{1st}, \textcolor[rgb]{0,0.5,0}{2nd}, \textcolor{blue}{3rd}, 4th)}%
  \label{tab:llff}
\vspace{-1.5em}
  \begin{tabular}{@{\hspace{0mm}} l @{\hspace{2mm}} c @{\hspace{2mm}} c @{\hspace{2mm}} c @{\hspace{2mm}} c @{\hspace{0mm}}}
  \hline
  Method  & JaxNeRF &  Plenoxels  & DVGO & TensoRF \\
  \hline
Fern & \textcolor{black}{24.83} / \textcolor{red}{21.58} & \textcolor{red}{25.47} / \textcolor{black}{19.07} & \textcolor{blue}{25.07} / \textcolor[rgb]{0,0.5,0}{20.04} & \textcolor[rgb]{0,0.5,0}{25.31} / \textcolor{blue}{19.68}\\
Flower & \textcolor[rgb]{0,0.5,0}{28.07} / \textcolor{red}{23.03} & \textcolor{blue}{27.83} / \textcolor{black}{19.46} & \textcolor{black}{27.61} / \textcolor[rgb]{0,0.5,0}{21.42} & \textcolor{red}{28.22} / \textcolor{blue}{21.15}\\
Fortress & \textcolor{red}{31.76} / \textcolor{red}{26.97} & \textcolor{blue}{31.09} / \textcolor{black}{22.32} & \textcolor{black}{30.38} / \textcolor[rgb]{0,0.5,0}{24.82} & \textcolor[rgb]{0,0.5,0}{31.14} / \textcolor{blue}{24.57}\\
Horns & \textcolor{red}{28.10} / \textcolor{red}{23.58} & \textcolor{blue}{27.60} / \textcolor{black}{19.09} & \textcolor{black}{27.55} / \textcolor{blue}{21.20} & \textcolor[rgb]{0,0.5,0}{27.64} / \textcolor[rgb]{0,0.5,0}{21.80}\\
Leaves & \textcolor{blue}{21.23} / \textcolor{red}{17.53} & \textcolor{red}{21.43} / \textcolor{black}{16.32} & \textcolor{black}{21.04} / \textcolor{blue}{16.43} & \textcolor[rgb]{0,0.5,0}{21.34} / \textcolor[rgb]{0,0.5,0}{16.58}\\
Orchids & \textcolor[rgb]{0,0.5,0}{20.27} / \textcolor{red}{18.06} & \textcolor{blue}{20.26} / \textcolor{black}{16.03} & \textcolor{red}{20.38} / \textcolor[rgb]{0,0.5,0}{16.59} & \textcolor{black}{20.02} / \textcolor{blue}{16.37}\\
Room & \textcolor{red}{33.04} / \textcolor{red}{26.08} & \textcolor{black}{30.22} / \textcolor{black}{21.64} & \textcolor{blue}{31.45} / \textcolor{blue}{24.64} & \textcolor[rgb]{0,0.5,0}{31.80} / \textcolor[rgb]{0,0.5,0}{25.19}\\
Trex & \textcolor{red}{27.42} / \textcolor{red}{24.22} & \textcolor{black}{26.49} / \textcolor{black}{19.23} & \textcolor[rgb]{0,0.5,0}{27.17} / \textcolor{blue}{20.90} & \textcolor{blue}{26.61} / \textcolor[rgb]{0,0.5,0}{21.23}\\
\hline
Average & \textcolor{red}{26.84} / \textcolor{red}{22.63} & \textcolor{black}{26.30} / \textcolor{black}{19.15} & \textcolor{blue}{26.33} / \textcolor{blue}{20.76} & \textcolor[rgb]{0,0.5,0}{26.51} / \textcolor[rgb]{0,0.5,0}{20.82}\\
\hline
  \end{tabular}
\end{center}
\vspace{-1em}
\end{table}

Benefiting from the IMRC metric, we could quantitatively but not only qualitatively evaluate how a regularization loss improves the density field. Specifically, we train DVGO~\cite{ChengSun2022DVGO} with distortion loss~\cite{barron2022mipnerf360}, which could improve the density field qualitatively by observations. The results are reported in \cref{tab:dis}. With distortion loss, the better IMRC for all the three datasets verified the effectiveness of both the distortion loss and the metric itself. 

\begin{table}[h]
\tiny
\begin{center}
\vspace{-0em}
  \caption{PSNR$\uparrow$/IMRC$\uparrow$ results on three datasets produced by DVGO with and without distortion loss. (\textbf{Better})}%
  \label{tab:dis}
\vspace{-1.5em}
  \begin{tabular}{@{\hspace{0mm}} l @{\hspace{2mm}} c @{\hspace{2mm}} c @{\hspace{2mm}} c @{\hspace{2mm}} c @{\hspace{0mm}}}
  \hline
  Dataset  & DTU &  NeRF Synthetic  & LLFF \\
  \hline
DVGO & 32.15 / 18.19 & \textbf{31.58} / 17.25 & 26.23 / 18.87\\
DVGO + Distortion Loss & \textbf{32.20} / \textbf{18.47} & 31.50 / \textbf{17.94} & \textbf{26.33} / \textbf{20.76} \\
\hline
  \end{tabular}
\end{center}
\vspace{-2em}
\end{table}

%

\section{Discussion and Conclusion}

We first discuss the limitations of IMRC and then conclude this paper in the following.

\textbf{Limitations}.
There are also some degrade situations where the proposed IMRC metric may fail.
For example, the low-frequency color prior no longer holds for surfaces that are purely specular or exhibit high levels of specular reflection, such as mirrors, leading to the failure of IMRC.
Another degrade situation is that, if the whole density field is empty except for a few points that have been observed only by several cameras, the IMRC metric will be high.
However, this density field is really bad.
This type of degrade situations also exist for other geometry metrics, such as the re-projection error.
Therefore, when using the proposed IMRC metric, it should be combined with other metrics, such as PSNR and SSIM.
Only in this way, we can evaluate the results with less bias.

\textbf{Conclusion}.
In this paper, we aim at evaluating the geometry information of a radiance field without ground-truth.
This problem is important as the radiance field has been widely used not only on novel view synthesis but also 3D reconstruction tasks.
For 3D reconstruction tasks, the geometry information is essential.
However, due to the unavailable ground-truth and unsuitable metrics, there is no proper metric to quantitatively evaluate the geometry.
To alleviate this dilemma, we propose the Inverse Mean Residual Color (IMRC) metric based on our insights on the properties of the radiance field.
Qualitative and quantitative experimental results verify the effectiveness of the IMRC metric.
We also benchmark $4$ state-of-the-art methods on $3$ datasets and hope this work could promote future related research.

\section{Acknowledgments}
This work was supported in part by the National Key Research and Development Program of China under Grant 2021YFB3301504, in part by the National Natural Science Foundation of China under Grant U19B2029, Grant U1909204, and Grant 92267103, in part by the Guangdong Basic and Applied Basic Research Foundation under Grant 2021B1515140034, and in part by the CAS STS Dongguan Joint Project under Grant 20211600200022.

\bibliography{aaai24,imrc_bib}

\begin{thebibliography}{54}
\providecommand{\natexlab}[1]{#1}

\bibitem[{Agarwal et~al.(2009)Agarwal, Snavely, Simon, Seitz, and Szeliski}]{Agarwal2009BuildingRome}
Agarwal, S.; Snavely, N.; Simon, I.; Seitz, S.~M.; and Szeliski, R. 2009.
\newblock {Building Rome in a Day}.
\newblock In \emph{ICCV}, 72--79.

\bibitem[{Barron et~al.(2021)Barron, Mildenhall, Tancik, Hedman, Martin-Brualla, and Srinivasan}]{barron2021MipNeRF}
Barron, J.~T.; Mildenhall, B.; Tancik, M.; Hedman, P.; Martin-Brualla, R.; and Srinivasan, P.~P. 2021.
\newblock Mip-nerf: A multiscale representation for anti-aliasing neural radiance fields.
\newblock In \emph{ICCV}, 5855--5864.

\bibitem[{Barron et~al.(2022)Barron, Mildenhall, Verbin, Srinivasan, and Hedman}]{barron2022mipnerf360}
Barron, J.~T.; Mildenhall, B.; Verbin, D.; Srinivasan, P.~P.; and Hedman, P. 2022.
\newblock Mip-nerf 360: Unbounded anti-aliased neural radiance fields.
\newblock In \emph{CVPR}, 5470--5479.

\bibitem[{Basri and Jacobs(2003)}]{Basri2003lambertianSH}
Basri, R.; and Jacobs, D.~W. 2003.
\newblock {Lambertian Reflectance and Linear Subspaces}.
\newblock \emph{IEEE TPAMI}, 25(2): 218--233.

\bibitem[{Chen et~al.(2022{\natexlab{a}})Chen, Xu, Geiger, Yu, and Su}]{AnpeiChen2022TensoRFTR}
Chen, A.; Xu, Z.; Geiger, A.; Yu, J.; and Su, H. 2022{\natexlab{a}}.
\newblock {TensoRF: Tensorial Radiance Fields}.
\newblock In \emph{ECCV}.

\bibitem[{Chen et~al.(2021)Chen, Xu, Zhao, Zhang, Xiang, Yu, and Su}]{chen2021mvsnerf}
Chen, A.; Xu, Z.; Zhao, F.; Zhang, X.; Xiang, F.; Yu, J.; and Su, H. 2021.
\newblock {MVSNeRF: Fast Generalizable Radiance Field Reconstruction From Multi-view Stereo}.
\newblock In \emph{ICCV}, 14124--14133.

\bibitem[{Chen et~al.(2022{\natexlab{b}})Chen, Funkhouser, Hedman, and Tagliasacchi}]{chen2022mobilenerf}
Chen, Z.; Funkhouser, T.; Hedman, P.; and Tagliasacchi, A. 2022{\natexlab{b}}.
\newblock {MobileNeRF: Exploiting the Polygon Rasterization Pipeline for Efficient Neural Field Rendering on Mobile Architectures}.
\newblock \emph{arXiv preprint arXiv:2208.00277}.

\bibitem[{Deng, Barron, and Srinivasan(2020)}]{jaxnerf2020github}
Deng, B.; Barron, J.~T.; and Srinivasan, P.~P. 2020.
\newblock {JaxNeRF}: an efficient {JAX} implementation of {NeRF}.

\bibitem[{Fang et~al.(2023)Fang, Song, Li, and Bo}]{fang2023reducing_nips23}
Fang, Q.; Song, Y.; Li, K.; and Bo, L. 2023.
\newblock {Reducing Shape-Radiance Ambiguity in Radiance Fields with a Closed-Form Color Estimation Method}.
\newblock In \emph{NIPS}.

\bibitem[{Garbin et~al.(2021)Garbin, Kowalski, Johnson, Shotton, and Valentin}]{garbin2021fastnerf}
Garbin, S.~J.; Kowalski, M.; Johnson, M.; Shotton, J.; and Valentin, J. 2021.
\newblock {FastNeRF: High-fidelity Neural Rendering at 200fps}.
\newblock In \emph{ICCV}, 14346--14355.

\bibitem[{Green(2003)}]{Green2003SH}
Green, R. 2003.
\newblock Spherical harmonic lighting: The gritty details.
\newblock In \emph{Archives of the game developers conference}, volume~56, 4.

\bibitem[{Hartley and Zisserman(2004)}]{Hartley2004MVG}
Hartley, R.~I.; and Zisserman, A. 2004.
\newblock \emph{{Multiple View Geometry in Computer Vision}}.
\newblock Cambridge University Press, ISBN: 0521540518, second edition.

\bibitem[{Hedman et~al.(2021)Hedman, Srinivasan, Mildenhall, Barron, and Debevec}]{PeterHedman2021BakingNR}
Hedman, P.; Srinivasan, P.~P.; Mildenhall, B.; Barron, J.~T.; and Debevec, P. 2021.
\newblock {Baking Neural Radiance Fields for Real-Time View Synthesis}.
\newblock In \emph{ICCV}.

\bibitem[{Hu et~al.(2022)Hu, Liu, Chen, Shen, and Jia}]{Hu2022EfficientNeRF}
Hu, T.; Liu, S.; Chen, Y.; Shen, T.; and Jia, J. 2022.
\newblock {EfficientNeRF - Efficient Neural Radiance Fields}.
\newblock In \emph{CVPR}, 12892--12901.

\bibitem[{Jain, Tancik, and Abbeel(2021)}]{jain2021putting}
Jain, A.; Tancik, M.; and Abbeel, P. 2021.
\newblock {Putting Nerf on A Diet: Semantically Consistent Few-shot View Synthesis}.
\newblock In \emph{ICCV}, 5885--5894.

\bibitem[{Jensen et~al.(2014)Jensen, Dahl, Vogiatzis, Tola, and Aan{\ae}s}]{Jensen2014DTU}
Jensen, R.; Dahl, A.; Vogiatzis, G.; Tola, E.; and Aan{\ae}s, H. 2014.
\newblock Large scale multi-view stereopsis evaluation.
\newblock In \emph{CVPR}, 406--413. IEEE.

\bibitem[{Li et~al.(2015)Li, Xia, Song, Fang, and Chen}]{li2015DataDriven_iccv15}
Li, J.; Xia, C.; Song, Y.; Fang, S.; and Chen, X. 2015.
\newblock A Data-Driven Metric for Comprehensive Evaluation of Saliency Models.
\newblock In \emph{ICCV}.

\bibitem[{Long et~al.(2022)Long, Lin, Wang, Komura, and Wang}]{Long2022SpareseNeuS}
Long, X.; Lin, C.; Wang, P.; Komura, T.; and Wang, W. 2022.
\newblock {SparseNeuS: Fast Generalizable Neural Surface Reconstruction from Sparse views}.
\newblock In \emph{ECCV}.

\bibitem[{Martin-Brualla et~al.(2020)Martin-Brualla, Radwan, Sajjadi, Barron, Dosovitskiy, and Duckworth}]{RicardoMartinBrualla2020NeRFIT}
Martin-Brualla, R.; Radwan, N.; Sajjadi, M. S.~M.; Barron, J.~T.; Dosovitskiy, A.; and Duckworth, D. 2020.
\newblock {NeRF in the Wild: Neural Radiance Fields for Unconstrained Photo Collections}.
\newblock In \emph{CVPR}.

\bibitem[{Max(1995)}]{max1995optical}
Max, N. 1995.
\newblock {Optical Models for Direct Volume Rendering}.
\newblock \emph{IEEE TVCG}, 1(2): 99--108.

\bibitem[{Mildenhall et~al.(2022)Mildenhall, Hedman, Martin-Brualla, Srinivasan, and Barron}]{mildenhall2022NeRFdark}
Mildenhall, B.; Hedman, P.; Martin-Brualla, R.; Srinivasan, P.~P.; and Barron, J.~T. 2022.
\newblock Nerf in the dark: High dynamic range view synthesis from noisy raw images.
\newblock In \emph{CVPR}, 16190--16199.

\bibitem[{Mildenhall et~al.(2019)Mildenhall, Srinivasan, Ortiz-Cayon, Kalantari, Ramamoorthi, Ng, and Kar}]{Mildenhall2019LLFF}
Mildenhall, B.; Srinivasan, P.~P.; Ortiz-Cayon, R.; Kalantari, N.~K.; Ramamoorthi, R.; Ng, R.; and Kar, A. 2019.
\newblock Local Light Field Fusion: Practical View Synthesis with Prescriptive Sampling Guidelines.
\newblock \emph{ACM Trans. Graph.}, 38(4).

\bibitem[{Mildenhall et~al.(2020)Mildenhall, Srinivasan, Tancik, Barron, Ramamoorthi, and Ng}]{BenMildenhall2020NeRFRS}
Mildenhall, B.; Srinivasan, P.~P.; Tancik, M.; Barron, J.~T.; Ramamoorthi, R.; and Ng, R. 2020.
\newblock {NeRF: Representing Scenes as Neural Radiance Fields for View Synthesis}.
\newblock In \emph{ECCV}.

\bibitem[{M{\"u}ller et~al.(2022)M{\"u}ller, Evans, Schied, and Keller}]{muller2022instantNGP}
M{\"u}ller, T.; Evans, A.; Schied, C.; and Keller, A. 2022.
\newblock {Instant Neural Graphics Primitives With A Multiresolution Hash Encoding}.
\newblock \emph{arXiv preprint arXiv:2201.05989}.

\bibitem[{Ng(2005)}]{ng2005fourier}
Ng, R. 2005.
\newblock Fourier slice photography.
\newblock In \emph{SIGGRAPH}, 735--744.

\bibitem[{Niemeyer et~al.(2022)Niemeyer, Barron, Mildenhall, Sajjadi, Geiger, and Radwan}]{Niemeyer2022RegNeRF}
Niemeyer, M.; Barron, J.~T.; Mildenhall, B.; Sajjadi, M. S.~M.; Geiger, A.; and Radwan, N. 2022.
\newblock {RegNeRF: Regularizing Neural Radiance Fields for View Synthesis from Sparse Inputs}.
\newblock In \emph{CVPR}.

\bibitem[{Oechsle, Peng, and Geiger(2021)}]{Oechsle2021UNISURF}
Oechsle, M.; Peng, S.; and Geiger, A. 2021.
\newblock {UNISURF: Unifying Neural Implicit Surfaces and Radiance Fields for Multi-View Reconstruction}.
\newblock In \emph{ICCV}, 5569--5579.

\bibitem[{Peng et~al.(2021)Peng, Dong, Wang, Zhang, Shuai, Zhou, and Bao}]{peng2021animatable}
Peng, S.; Dong, J.; Wang, Q.; Zhang, S.; Shuai, Q.; Zhou, X.; and Bao, H. 2021.
\newblock {Animatable Neural Radiance Fields for Modeling Dynamic Human Bodies}.
\newblock In \emph{ICCV}, 14314--14323.

\bibitem[{Ramamoorthi and Hanrahan(2001)}]{Ramamoorthi2001relationshipSH}
Ramamoorthi, R.; and Hanrahan, P. 2001.
\newblock On the Relationship Between Radiance and Irradiance: Determining the Illumination From Images of a Convex Lambertian Object.
\newblock \emph{JOSA A}, 18(10): 2448--2459.

\bibitem[{Reiser et~al.(2021)Reiser, Peng, Liao, and Geiger}]{Reiser2021KiloNeRF}
Reiser, C.; Peng, S.; Liao, Y.; and Geiger, A. 2021.
\newblock {KiloNeRF: Speeding up Neural Radiance Fields with Thousands of Tiny MLPs}.
\newblock In \emph{ICCV}.

\bibitem[{Shao et~al.(2022)Shao, Zhang, Zhang, Chen, Cao, Yu, and Liu}]{Shao2022DoubleField}
Shao, R.; Zhang, H.; Zhang, H.; Chen, M.; Cao, Y.; Yu, T.; and Liu, Y. 2022.
\newblock {DoubleField: Bridging the Neural Surface and Radiance Fields for High-fidelity Human Reconstruction and Rendering}.
\newblock In \emph{CVPR}.

\bibitem[{Snavely, Seitz, and Szeliski(2006)}]{Snavely2006PhotoTourism}
Snavely, N.; Seitz, S.~M.; and Szeliski, R. 2006.
\newblock Photo Tourism: Exploring Photo Collections in 3D.
\newblock \emph{ACM TOG}, 25(3): 835–846.

\bibitem[{Snavely, Seitz, and Szeliski(2008)}]{Snavely2008Modeling}
Snavely, N.; Seitz, S.~M.; and Szeliski, R. 2008.
\newblock Modeling the World from Internet Photo Collections.
\newblock \emph{IJCV}, 80(2): 189--210.

\bibitem[{Sun, Sun, and Chen(2022{\natexlab{a}})}]{ChengSun2022DVGO}
Sun, C.; Sun, M.; and Chen, H.-T. 2022{\natexlab{a}}.
\newblock {Direct Voxel Grid Optimization: Super-fast Convergence for Radiance Fields Reconstruction}.
\newblock In \emph{CVPR}.

\bibitem[{Sun, Sun, and Chen(2022{\natexlab{b}})}]{sun2022DVGOv2}
Sun, C.; Sun, M.; and Chen, H.-T. 2022{\natexlab{b}}.
\newblock {Improved Direct Voxel Grid Optimization for Radiance Fields Reconstruction}.
\newblock \emph{arXiv preprint arXiv:2206.05085}.

\bibitem[{Sun et~al.(2022)Sun, Chen, Wang, Li, Averbuch-Elor, Zhou, and Snavely}]{Sun2022SIGGRAPH}
Sun, J.; Chen, X.; Wang, Q.; Li, Z.; Averbuch-Elor, H.; Zhou, X.; and Snavely, N. 2022.
\newblock {Neural 3D Reconstruction in the Wild}.
\newblock In \emph{SIGGRAPH}.

\bibitem[{Tancik et~al.(2022)Tancik, Casser, Yan, Pradhan, Mildenhall, Srinivasan, Barron, and Kretzschmar}]{Tancik2022BlockNeRF}
Tancik, M.; Casser, V.; Yan, X.; Pradhan, S.; Mildenhall, B.; Srinivasan, P.~P.; Barron, J.~T.; and Kretzschmar, H. 2022.
\newblock {Block-NeRF: Scalable Large Scene Neural View Synthesis}.
\newblock In \emph{CVPR}.

\bibitem[{Wang et~al.(2023{\natexlab{a}})Wang, Chen, Wang, Song, and Liu}]{wang2023masked_nips23}
Wang, F.; Chen, Z.; Wang, G.; Song, Y.; and Liu, H. 2023{\natexlab{a}}.
\newblock {Masked Space-Time Hash Encoding for Efficient Dynamic Scene Reconstruction}.
\newblock In \emph{NIPS}.

\bibitem[{Wang et~al.(2023{\natexlab{b}})Wang, Tan, Li, Tian, Song, and Liu}]{wang2023mixed_iccv23}
Wang, F.; Tan, S.; Li, X.; Tian, Z.; Song, Y.; and Liu, H. 2023{\natexlab{b}}.
\newblock {Mixed Neural Voxels for Fast Multi-view Video Synthesis}.
\newblock In \emph{ICCV}, 19706--19716.

\bibitem[{Wang et~al.(2021{\natexlab{a}})Wang, Liu, Liu, Theobalt, Komura, and Wang}]{Wang2021NeuS}
Wang, P.; Liu, L.; Liu, Y.; Theobalt, C.; Komura, T.; and Wang, W. 2021{\natexlab{a}}.
\newblock {NeuS: Learning Neural Implicit Surfaces by Volume Rendering for Multi-view Reconstruction}.
\newblock In \emph{NeurIPS}.

\bibitem[{Wang et~al.(2021{\natexlab{b}})Wang, Wang, Genova, Srinivasan, Zhou, Barron, Martin-Brualla, Snavely, and Funkhouser}]{wang2021IBRNet}
Wang, Q.; Wang, Z.; Genova, K.; Srinivasan, P.~P.; Zhou, H.; Barron, J.~T.; Martin-Brualla, R.; Snavely, N.; and Funkhouser, T. 2021{\natexlab{b}}.
\newblock {IBRNet: Learning Multi-view Image-based Rendering}.
\newblock In \emph{CVPR}, 4690--4699.

\bibitem[{Wang et~al.(2022)Wang, Han, Habermann, Daniilidis, Theobalt, and Liu}]{Wang2022InstanNGP_NeuS}
Wang, Y.; Han, Q.; Habermann, M.; Daniilidis, K.; Theobalt, C.; and Liu, L. 2022.
\newblock {NeuS2: Fast Learning of Neural Implicit Surfaces for Multi-view Reconstruction}.
\newblock \emph{arXiv}.

\bibitem[{Weng et~al.(2022)Weng, Curless, Srinivasan, Barron, and Kemelmacher-Shlizerman}]{Weng2022HumanNeRFMono}
Weng, C.-Y.; Curless, B.; Srinivasan, P.~P.; Barron, J.~T.; and Kemelmacher-Shlizerman, I. 2022.
\newblock {HumanNeRF: Free-viewpoint Rendering of Moving People from Monocular Video}.
\newblock In \emph{CVPR}.

\bibitem[{Wu et~al.(2022{\natexlab{a}})Wu, Liu, Chen, Li, Zheng, Cai, and Zheng}]{wu2022ObjectSDF}
Wu, Q.; Liu, X.; Chen, Y.; Li, K.; Zheng, C.; Cai, J.; and Zheng, J. 2022{\natexlab{a}}.
\newblock {Object-Compositional Neural Implicit Surfaces}.
\newblock In \emph{ECCV}.

\bibitem[{Wu et~al.(2023)Wu, Wang, Li, Zheng, and Cai}]{Wu2023OjbectSDF++}
Wu, Q.; Wang, K.; Li, K.; Zheng, J.; and Cai, J. 2023.
\newblock {ObjectSDF++: Improved Object-Compositional Neural Implicit Surfaces}.
\newblock In \emph{ICCV}.

\bibitem[{Wu et~al.(2022{\natexlab{b}})Wu, Wang, Pan, Xu, Theobalt, Liu, and Lin}]{wu2022voxurf}
Wu, T.; Wang, J.; Pan, X.; Xu, X.; Theobalt, C.; Liu, Z.; and Lin, D. 2022{\natexlab{b}}.
\newblock Voxurf: Voxel-based Efficient and Accurate Neural Surface Reconstruction.
\newblock arXiv:2208.12697.

\bibitem[{Yariv et~al.(2020)Yariv, Kasten, Moran, Galun, Atzmon, Ronen, and Lipman}]{Yariv2020IDR}
Yariv, L.; Kasten, Y.; Moran, D.; Galun, M.; Atzmon, M.; Ronen, B.; and Lipman, Y. 2020.
\newblock Multiview Neural Surface Reconstruction by Disentangling Geometry and Appearance.
\newblock In Larochelle, H.; Ranzato, M.; Hadsell, R.; Balcan, M.; and Lin, H., eds., \emph{NeurIPS}, volume~33, 2492--2502. Curran Associates, Inc.

\bibitem[{Yu et~al.(2022)Yu, Fridovich-Keil, Tancik, Chen, Recht, and Kanazawa}]{AlexYu2022PlenoxelsRF}
Yu, A.; Fridovich-Keil, S.; Tancik, M.; Chen, Q.; Recht, B.; and Kanazawa, A. 2022.
\newblock {Plenoxels: Radiance Fields Without Neural Networks}.
\newblock In \emph{CVPR}.

\bibitem[{Yu et~al.(2021{\natexlab{a}})Yu, Li, Tancik, Li, Ng, and Kanazawa}]{AlexYu2021PlenOctreesFR}
Yu, A.; Li, R.; Tancik, M.; Li, H.; Ng, R.; and Kanazawa, A. 2021{\natexlab{a}}.
\newblock {PlenOctrees for Real-time Rendering of Neural Radiance Fields}.
\newblock In \emph{ICCV}.

\bibitem[{Yu et~al.(2021{\natexlab{b}})Yu, Ye, Tancik, and Kanazawa}]{yu2021pixelnerf}
Yu, A.; Ye, V.; Tancik, M.; and Kanazawa, A. 2021{\natexlab{b}}.
\newblock {PixelNeRF: Neural Radiance Fields From One or Few Images}.
\newblock In \emph{CVPR}, 4578--4587.

\bibitem[{Zhang et~al.(2021)Zhang, Yang, Tulsiani, and Ramanan}]{Zhang2021NeRS}
Zhang, J.~Y.; Yang, G.; Tulsiani, S.; and Ramanan, D. 2021.
\newblock {NeRS: Neural Reflectance Surfaces for Sparse-view 3D Reconstruction in the Wild}.
\newblock In \emph{NeurIPS}.

\bibitem[{Zhang et~al.(2022)Zhang, Luan, Li, and Snavely}]{Zhang2022IRON}
Zhang, K.; Luan, F.; Li, Z.; and Snavely, N. 2022.
\newblock {IRON: Inverse Rendering by Optimizing Neural SDFs and Materials from Photometric Images}.
\newblock In \emph{CVPR}.

\bibitem[{Zhang et~al.(2020)Zhang, Riegler, Snavely, and Koltun}]{KaiZhang2020NeRF++AA}
Zhang, K.; Riegler, G.; Snavely, N.; and Koltun, V. 2020.
\newblock {NeRF++: Analyzing and Improving Neural Radiance Fields}.
\newblock \emph{arXiv preprint arXiv:2010.07492}.

\bibitem[{Zhao et~al.(2022)Zhao, Yang, Zhang, Lin, Zhang, Yu, and Xu}]{Zhao2022HumanNeRFSparse}
Zhao, F.; Yang, W.; Zhang, J.; Lin, P.; Zhang, Y.; Yu, J.; and Xu, L. 2022.
\newblock {HumanNeRF: Efficiently Generated Human Radiance Field from Sparse Inputs}.
\newblock In \emph{CVPR}.

\end{thebibliography}

\clearpage
\onecolumn
\appendix

\section*{Appendix}

\section{More Experimental Details}

\subsection{Search for the Best Chamfer Distance}
\label{sec:CD_search}

To calculate the Chamfer Distance (CD) metric on the DTU dataset, we first transform the density volume into a point-cloud using the marching cubes algorithm.
The algorithm searches for iso-surfaces from the volume, where a hyper-parameter about the density level needs to be manually set.
The hyper-parameter usually significantly affects the resulting CD value. 
Therefore, we vary this hyper-parameter and search for the best CD. 
Specifically, we adopt the golden section search algorithm, which effectively finds the local minimum, or probably the global minimum, of the CD in our case.
The searching process is ended when the distance between the last two searched CD values is no greater than $0.001$. 

\begin{figure}[hpt]
  \centering
   \includegraphics[width=1.0\linewidth]{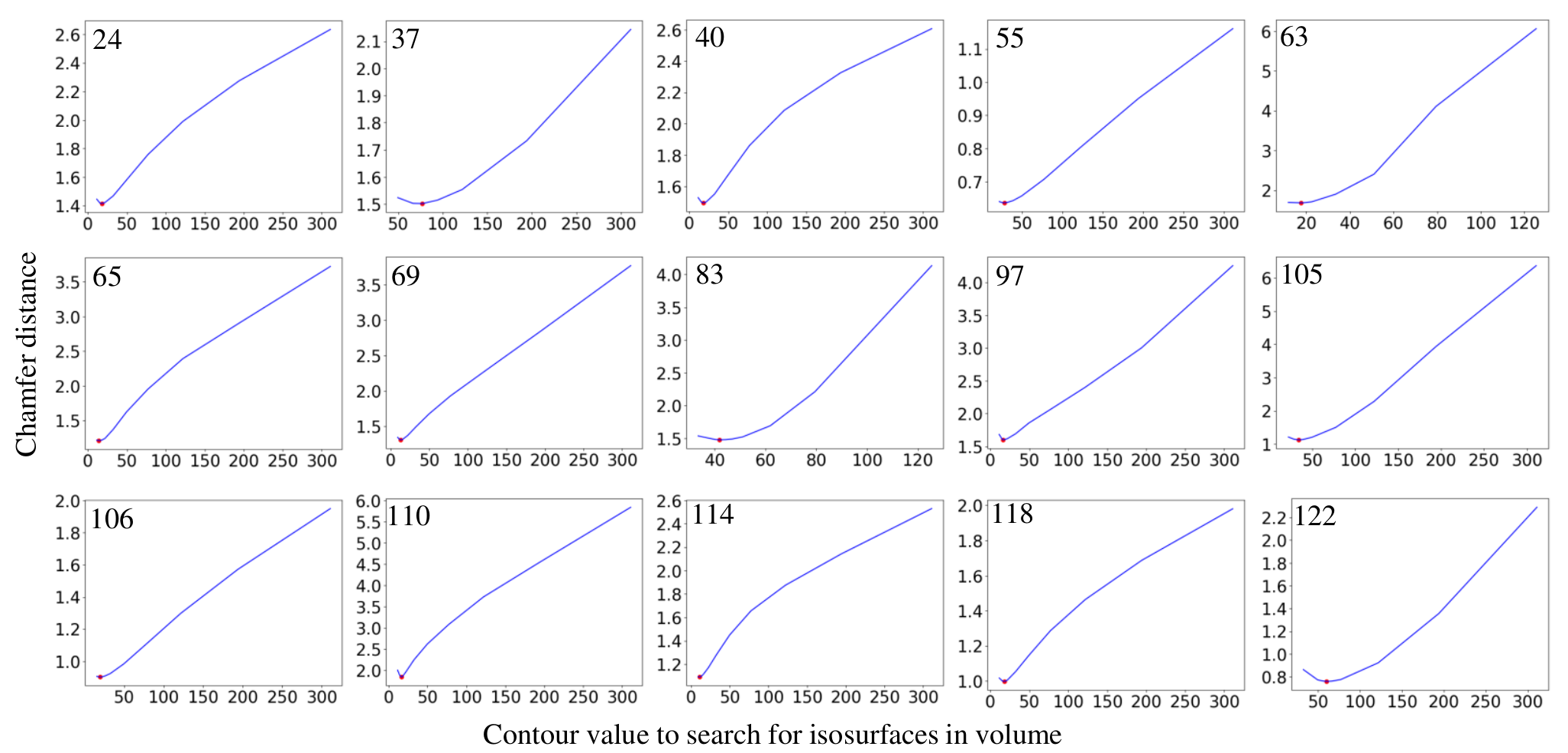}

   \caption{The searched results of the Chamfer Distance (CD) for each scene reconstructed by JaxNeRF~\cite{BenMildenhall2020NeRFRS, jaxnerf2020github}. The red points indicate the best CD values. The scan ID of the scene is listed on the left-top corner of each sub-figure.}
   \label{fig:jaxnerf_cd}
\end{figure}

As in \cref{fig:jaxnerf_cd}, we showcase the searched results of each scene reconstructed by JaxNeRF \cite{BenMildenhall2020NeRFRS, jaxnerf2020github}, which are based on $512^3$ density volume. The searched results first decrease and then increase. The ones of other methods, \textit{i.e.}, DVGO \cite{ChengSun2022DVGO}, Plenoxels \cite{AlexYu2022PlenoxelsRF}, and TensoRF \cite{AnpeiChen2022TensoRFTR}, have the similar trend. It is noteworthy that such a searching process is time-consuming, which costs tens of minutes or more, whereas the calculation of our Inverse Mean Residual Color (IMRC) metric is more efficient as reported in the next subsection.

\subsection{Computational Cost of the IMRC}
\label{sec:time}

As in \cref{tab:method_time}, we report the average time cost of calculating the IMRC metric on the DTU dataset’s 15 scenes with density volume resolution $512^3$ and SH degree $2$. 
All the experiments are run on a single NVIDIA A100 GPU. 
The average computation costs of the $4$ methods are different since the sparsity of the density volume varies with different methods.
More sparse the density volume is, less time the computation process costs.
Compared with the CD metric, the calculation of IMRC is much more efficient as there is no searching process.

\begin{table}[ht]
\small
\begin{center}
  \caption{Average time cost of calculating the IMRC on the DTU dataset.}%
  \label{tab:method_time}
  \begin{tabular}{@{\hspace{2mm}} l @{\hspace{5mm}} c @{\hspace{5mm}} c @{\hspace{5mm}} c @{\hspace{5mm}} c @{\hspace{2mm}}}
  \toprule
  Method  & JaxNeRF  &  Plenoxels  & DVGO & TensoRF \\
  \midrule
  Time cost (seconds)     & 17.80 & 16.98 & 10.66 & 16.50  \\
  \bottomrule
  \end{tabular}
\end{center}
\end{table}

\section{More Experimental Results}
\label{sec:result}

\subsection{Further Validation of IMRC by Depth Supervision and Comparison with Other Geometry Metrics}
\subsubsection{Depth Supervison}
To provide empirical evidence that the IMRC metric aligns with ground-truth density fields, a comparison between NeRF models trained with and without ground-truth depth supervision is conducted. It is expected that the models trained with depth supervision have better IMRC results.  
Specifically, we conduct an experiment on the NeRF synthetic dataset. Since only the 200 test images of each scene have the depth ground-truth, we further evenly split them into 2 subsets for training and testing. The results of DVGO trained with and without depth supervision are reported in Tab.~\ref{tab:with_geometric}. With depth supervision, the scenes should have better geometry. Accordingly, Tab.~\ref{tab:with_geometric} shows that the IMRC (the higher the better) of all scenes increases, which means that it aligns with ground-truth density fields.

\begin{table}[ht]
\begin{center}
\caption{PSNR, IMRC, mean squared error (MSE) of depth images, and mean angular error (MAE) results w/ and w/o depth supervision. (\textbf{Better})}%
\label{tab:with_geometric}

\begin{tabular}{@{\hspace{1mm}}l@{\hspace{3mm}}c@{\hspace{2mm}} c @{\hspace{2mm}}c@{\hspace{2mm}}c@{\hspace{4mm}} c@{\hspace{2mm}}c@{\hspace{2mm}}c@{\hspace{2mm}}c@{\hspace{1mm}}}
\hline
          & \multicolumn{4}{c}{With depth supervision}      & \multicolumn{4}{c}{Without depth supervision}  \\
          & PSNR$\uparrow$  & IMRC$\uparrow$   & MSE$\downarrow$  & MAE\degree$\downarrow$ & PSNR$\uparrow$ & IMRC$\uparrow$ & MSE$\downarrow$ & MAE\degree$\downarrow$\\
\hline
Chair     & 34.88             & \textbf{21.77}   & \textbf{0.0188}  & \textbf{55.47}  & \textbf{34.89} & 21.62 & 0.0196  & 55.72 \\
Drums     & \textbf{28.42}    & \textbf{14.86}   & \textbf{0.0786}  & \textbf{64.58}  & 28.40          & 13.82 & 0.1153  & 67.08 \\
Ficus     & \textbf{37.01}    & \textbf{19.34}   & \textbf{0.1139}  & \textbf{75.05} & 37.00          & 18.96 & 0.1501  & 75.38 \\
Hotdog    & \textbf{39.25}    & \textbf{23.34}   & \textbf{0.0125}  & \textbf{47.40} & 39.16          & 22.15 & 0.0259  & 48.54 \\
Lego      & 37.92             & \textbf{21.00}   & \textbf{0.0253}  & \textbf{60.17} & \textbf{37.95} & 19.83 & 0.0785  & 61.16 \\
Materials & \textbf{35.12}    & \textbf{17.40}   & \textbf{0.0334}  & \textbf{67.03} & 35.11          & 17.37 & 0.0394  & 68.45 \\
Mic       & \textbf{35.35}    & \textbf{18.38}   & \textbf{0.0507}  & \textbf{61.26} & 29.12          & 10.34 & 0.3760  & 69.72 \\
Ship      & \textbf{31.52}    & \textbf{19.85}   & \textbf{0.0531}  & \textbf{64.18} & 31.21          & 19.59 & 0.0611  & 64.79 \\
\hline
Average   & \textbf{34.93} & \textbf{19.49} & \textbf{0.0483} & \textbf{61.89} & 34.11         & 17.96 & 0.1082 & 63.86\\
\hline
\end{tabular}
\end{center}
\end{table}

\subsubsection{Comparison with Other Geometry Metrics}
To bolster the metric's credibility, we also demonstrate its correlation with commonly used geometry metrics, including the mean squared error (MSE) of depth images and the mean angular error (MAE) of normals, in Tab.~\ref{tab:with_geometric}. It shows that the variation tendency of IMRC is consistent with MSE and MAE. Moreover, IMRC does not need any geometric ground-truth which is hard to obtain especially for real-world data. Besides the synthetic data, for real-world data, the comparisons with the CD metric on the DTU dataset (please refer to Tab.~1 in the paper) also verify the metric's credibility.

\subsection{Scene-independent Nature of IMRC}
A desirable metric should be independent of the scenes to be evaluated. For instance, it is better that the best values of a metric for different scenes are the same. Given the ground-truth density fields of different scenes, their IMRC can be different. However, IMRC may not vary significantly. Due to the absence of ground-truth density fields, we resort to the ``practical'' upper-bound of each scene, which is defined as the best IMRC of a scene obtained by all currently available methods. In this way, the biases of different methods are eliminated as far as possible. In Tab.~\ref{tab:imrc_upperbound}, we present the average, standard deviation (SD), and coefficient of variation (CV), defined as SD / Average, of IMRC's practical upper-bounds across scenes. For comparison, we also present the corresponding results of PSNR. We can see that, on all the 3 datasets, all the SDs of IMRC are lower than those of PSNR. And only on the NeRF synthetic dataset, IMRC has a higher CV than PSNR,  since IMRC has an obviously lower average value. This experiment demonstrates that the variation of IMRC's practical upper-bounds across different scenes is acceptable compared with the common metric PSNR. In other words, scene-dependent nature has a limited influence on IMRC.

\begin{table}[ht]
\begin{center}
\caption{The average, standard deviation (SD), and coefficient of variation (CV) of IMRC's and PSNR's practical upper-bounds on 3 datasets.}%
\label{tab:imrc_upperbound}
\begin{tabular}{lcccccc}
\hline
               & \multicolumn{3}{c}{Practical upper-bound of IMRC} & \multicolumn{3}{c}{Practical upper-bound of PSNR} \\
               & Average   & SD $\downarrow$     & CV$\downarrow$       & Average   & SD$\downarrow$     & CV$\downarrow$       \\
\hline
DTU            & 19.07     & \textbf{1.92}   & \textbf{10.07}\%  & 32.50     & 3.35   & 10.31\%  \\
NeRF Synthetic & 18.59     & \textbf{2.65}   & 14.25\%  & 32.14     & 3.51   & \textbf{10.92}\%  \\
LLFF           & 22.63     & \textbf{3.43}   & \textbf{15.16}\%  & 26.98     & 4.46   & 16.53\%  \\
\hline
Mean           & 20.10     & \textbf{2.67}   & 13.16\%  & 30.54     & 3.77   & \textbf{12.59}\% \\
\hline
\end{tabular}
\end{center}
\end{table}

\subsection{Analysis on Reflective Objects and Typical Geometry Artifacts}

\subsubsection{Reflective Object}

\begin{figure}[ht]
  \centering
   \includegraphics[width=0.6\linewidth]{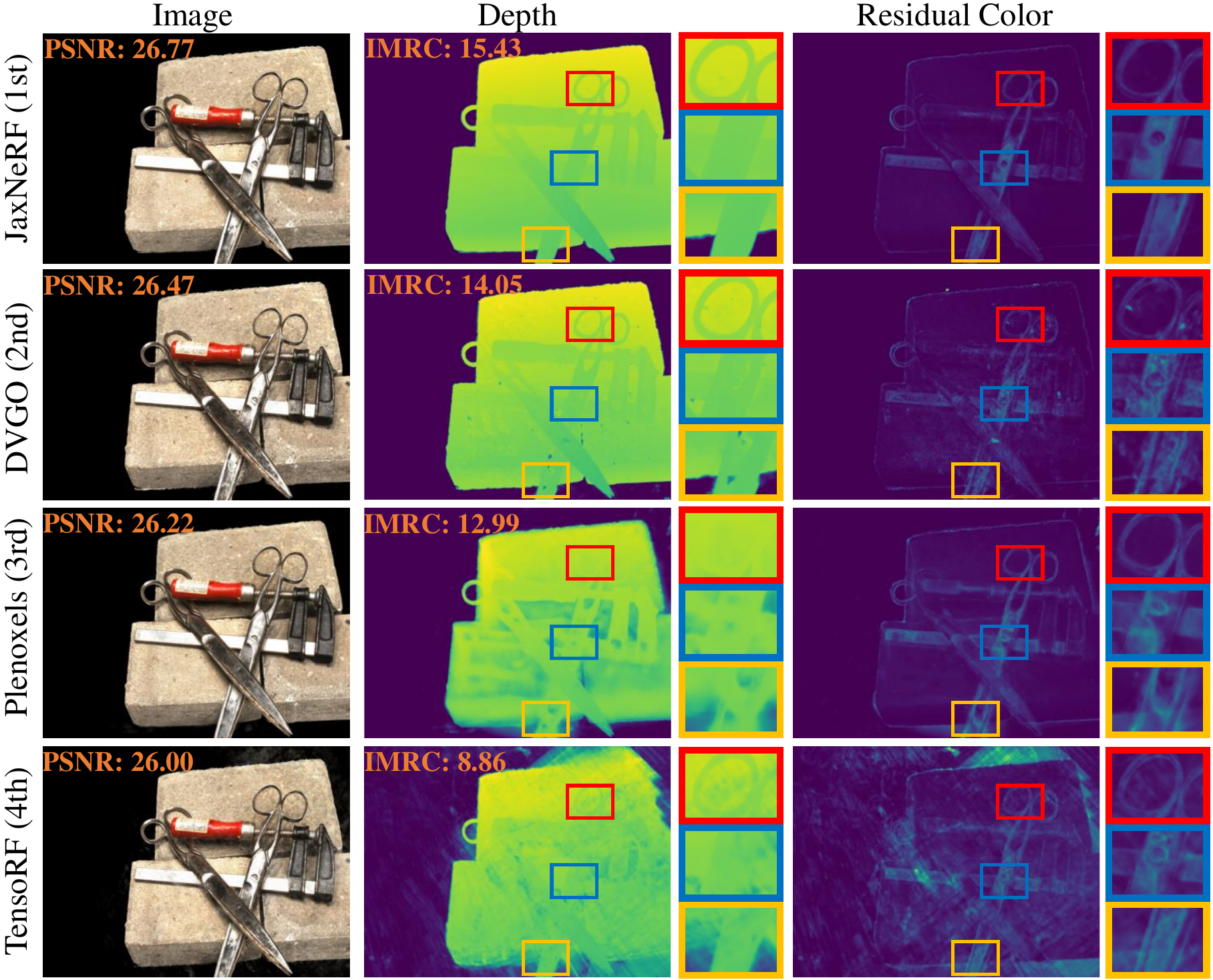}
   \caption{A reflective scene (Scan 37) in the DTU dataset. The UserRank becomes worse from top to down.}
   \label{fig:reflective}
\end{figure}

The IMRC metric is based on the low-frequency color prior. We use SH of degree $2$ to approximate the view-dependent colors from different observation directions. The degree $2$, rather than $0$, ensures that we can deal with non-Lambertian surfaces. The higher the SH degree, the better the non-Lambertian surfaces be approximated. With degree $2$, the highly reflective surface may have a high residual color. However, we are not intended to get a perfect approximation, which is not necessary. \textbf{The IMRC makes sense if it can correctly rank the geometry of the same scene produced by different methods}.
We demonstrate this by a reflective object as shown in \cref{fig:reflective}. The surface of the scissors is highly reflective, which leads to relatively high residual colors for a visually good geometry produced by JaxNeRF. Notice that because degree $2$ is applied to all methods, it is a fair treatment that all method has high residual colors on such a surface. More importantly, if one method produces worse geometry than that of JaxNeRF, its residual color will be much higher. This makes IMRC successfully distinguish and rank scene geometry even for a reflective case. 

\subsubsection{Thick Surface}

\begin{figure}[ht]
  \centering
   \includegraphics[width=0.9\linewidth]{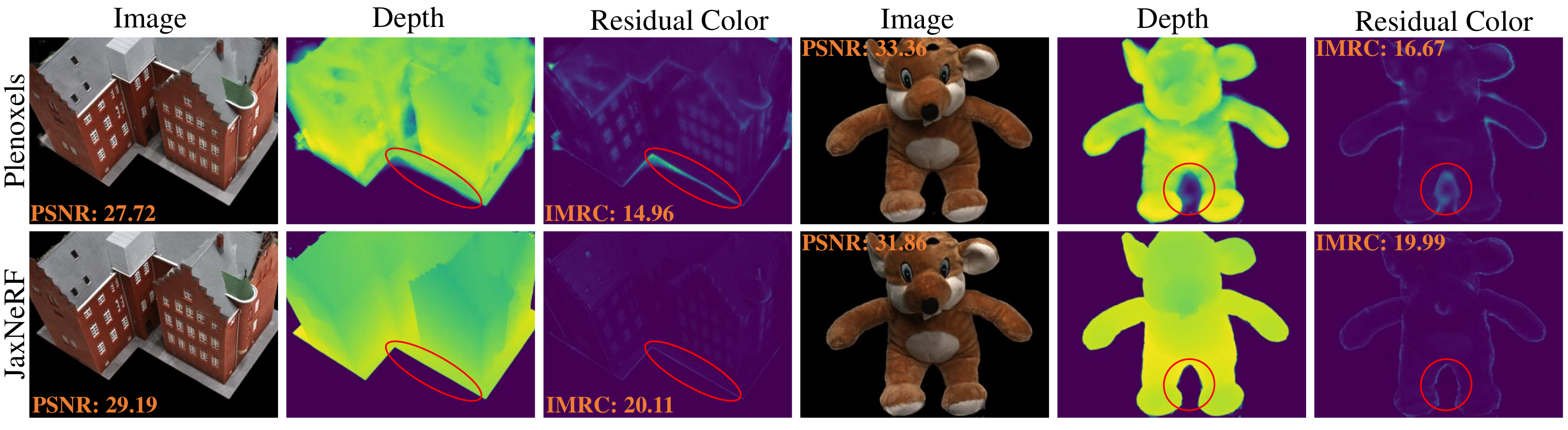}
   \caption{Two cases of thick surfaces (top row) on the DTU dataset. From left to right, Scan 24 and Scan 105. Corresponding sharp surface results (bottom row) are also presented for comparison.}
   \label{fig:thick}
\end{figure}

The thick surface is a typical geometry artifact that would generate inaccurate disparity maps viewed from different locations. Because CD is only based on an iso-surface at a typical density level, it may not be aware about the thickness of a surface. In contrast, IMRC can well recognize such an artifact, because points on the thick surface always have higher residual colors than those that lie nearer to the true surface. We showcase two cases that the IMRC metric penalties thick surfaces in \cref{fig:thick}. A sharper surface is also presented for comparison.

\subsubsection{Floating Surface}

\begin{figure}[ht]
  \centering
   \includegraphics[width=0.9\linewidth]{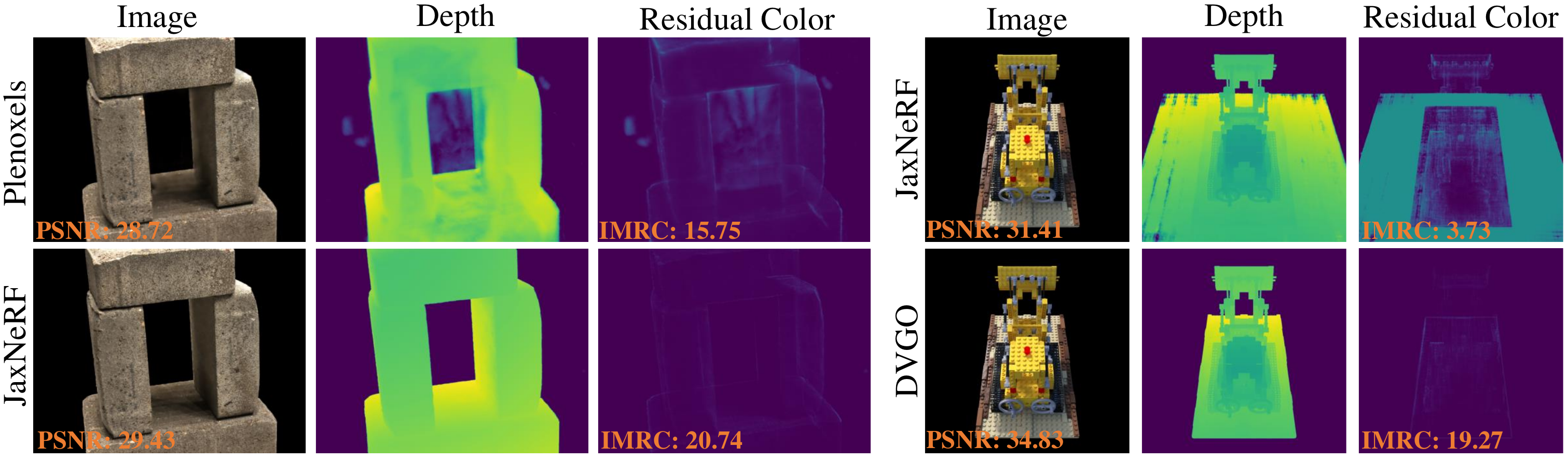}
   \caption{Two cases of floating surfaces (top row). From left to right, Scan 40 in the DTU dataset and Lego in the NeRF Synthetic dataset. Corresponding clean surface results (bottom row) are also presented for comparison.}
   \label{fig:float}
\end{figure}

The floating surface is a common artifact in NeRF models. The floaters violate the low-frequency color prior, and so will have high residual colors. We showcase two cases that the IMRC metric penalties floating surfaces in \cref{fig:float}. The better geometry without floating surfaces is also presented for comparison.  

\subsection{Remaining IMRC/UserRank and CD/UserRank Conflict Results}

\begin{figure}[ht]
  \centering
   \includegraphics[width=0.7\linewidth]{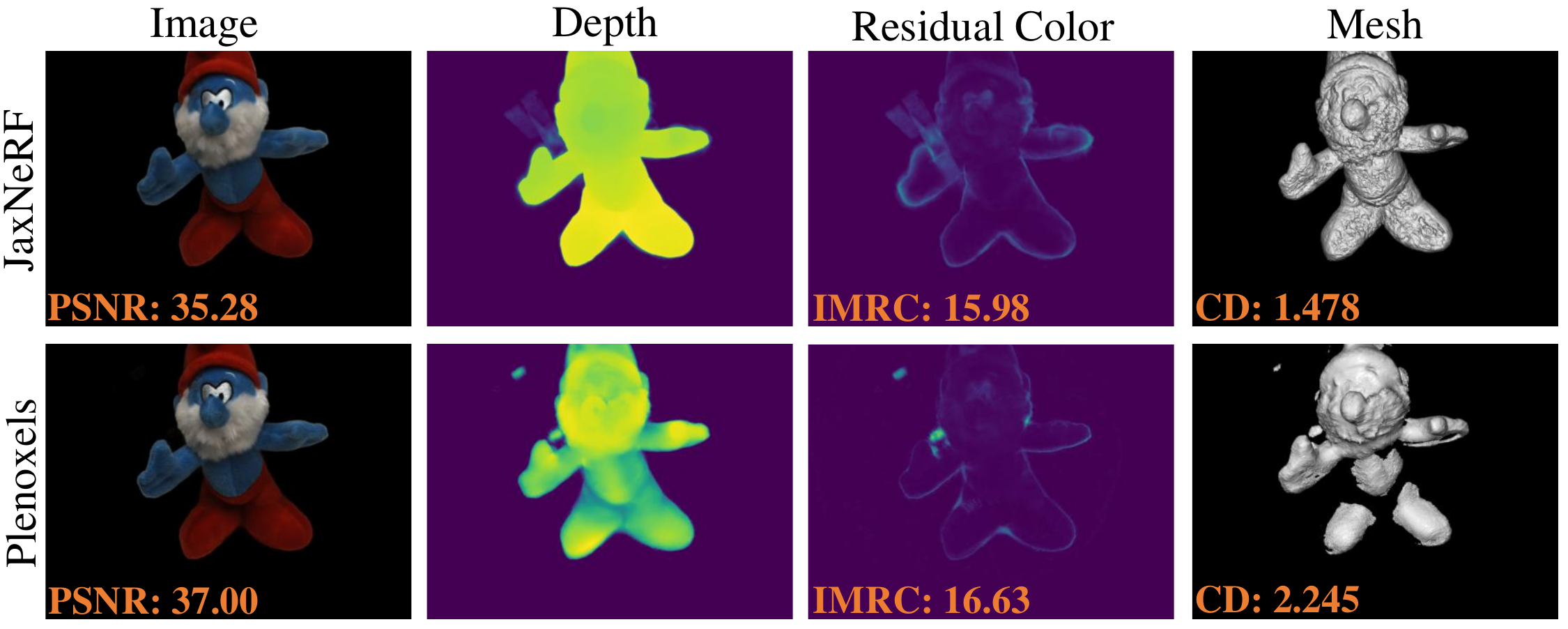}

   \caption{One remaining IMRC$\uparrow$/UserRank conflict case on the Scan 83 scene. The top row has a better UserRank.}
   \label{fig:imrc_user}
\end{figure}

We showcase one remaining IMRC$\uparrow$/UserRank conflict case in \cref{fig:imrc_user}. On the whole, JaxNeRF produces a thicker surface than Plenoxels as its residual color is always higher surrounding the contour. The Plenoxels produces some floating surfaces and perform worse on certain local details. Overall, the density fields of both methods are not good. After applying the marching cubes algorithm,  the floating surface of JaxNeRF's density field near the right hand of the doll is missing. In contrast, some low density parts of Plenoxels' density field are discarded, resulting in a poor mesh. Comprehensively considering the depth, residual color, and mesh, the users rank JaxNeRF as better, although the calculation results show that the Plenoxels has a lower residual color, or higher IMRC.

\begin{figure}[ht]
  \centering
   \includegraphics[width=1\linewidth]{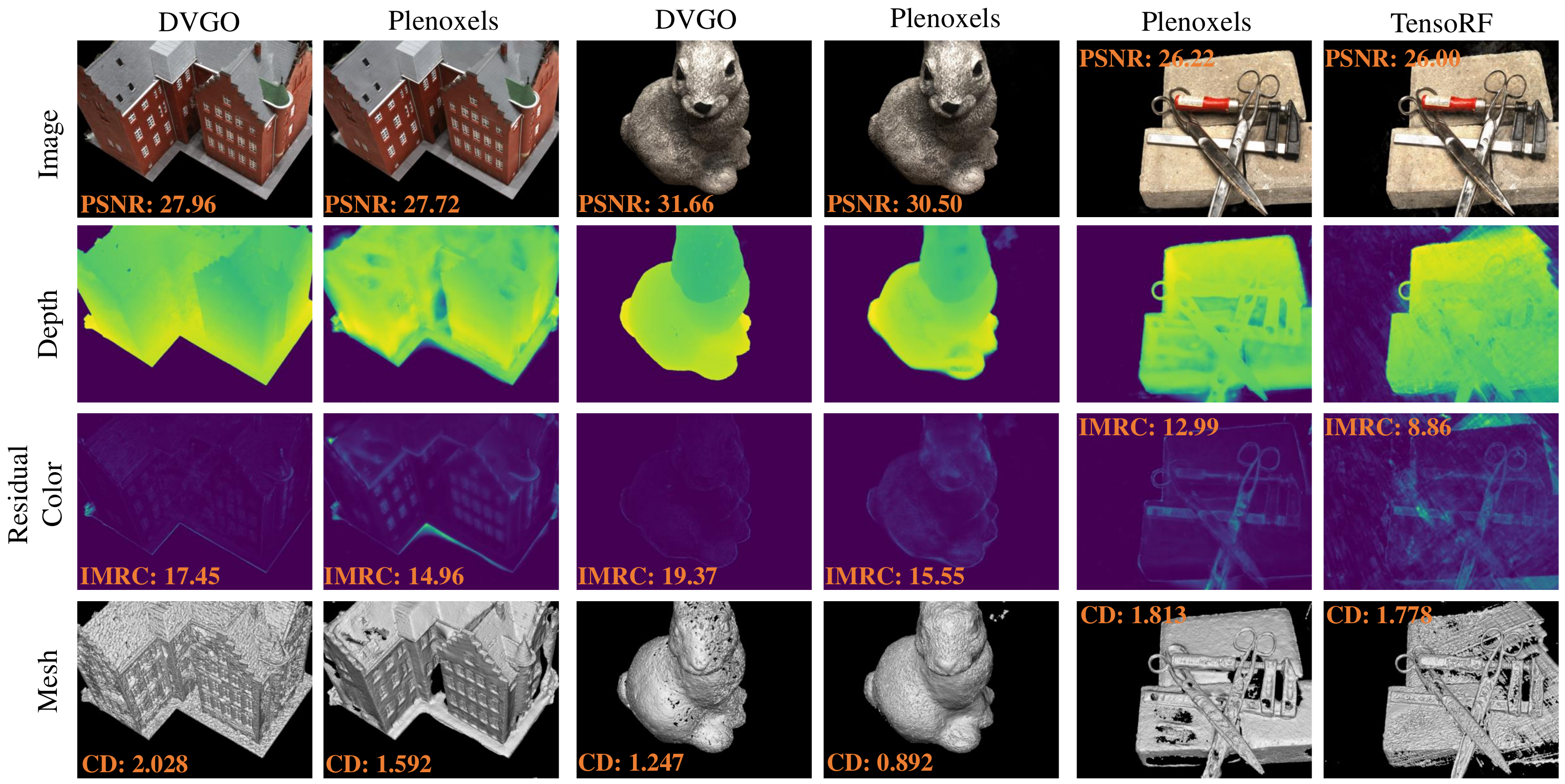}

   \caption{Three CD$\downarrow$/UserRank conflict cases. From left to right, DTU Scan 24, 55, and 37. For each scene, left has a better UserRank.}
   \label{fig:cd_user_1}
\end{figure}

\begin{figure}[ht]
  \centering
   \includegraphics[width=1\linewidth]{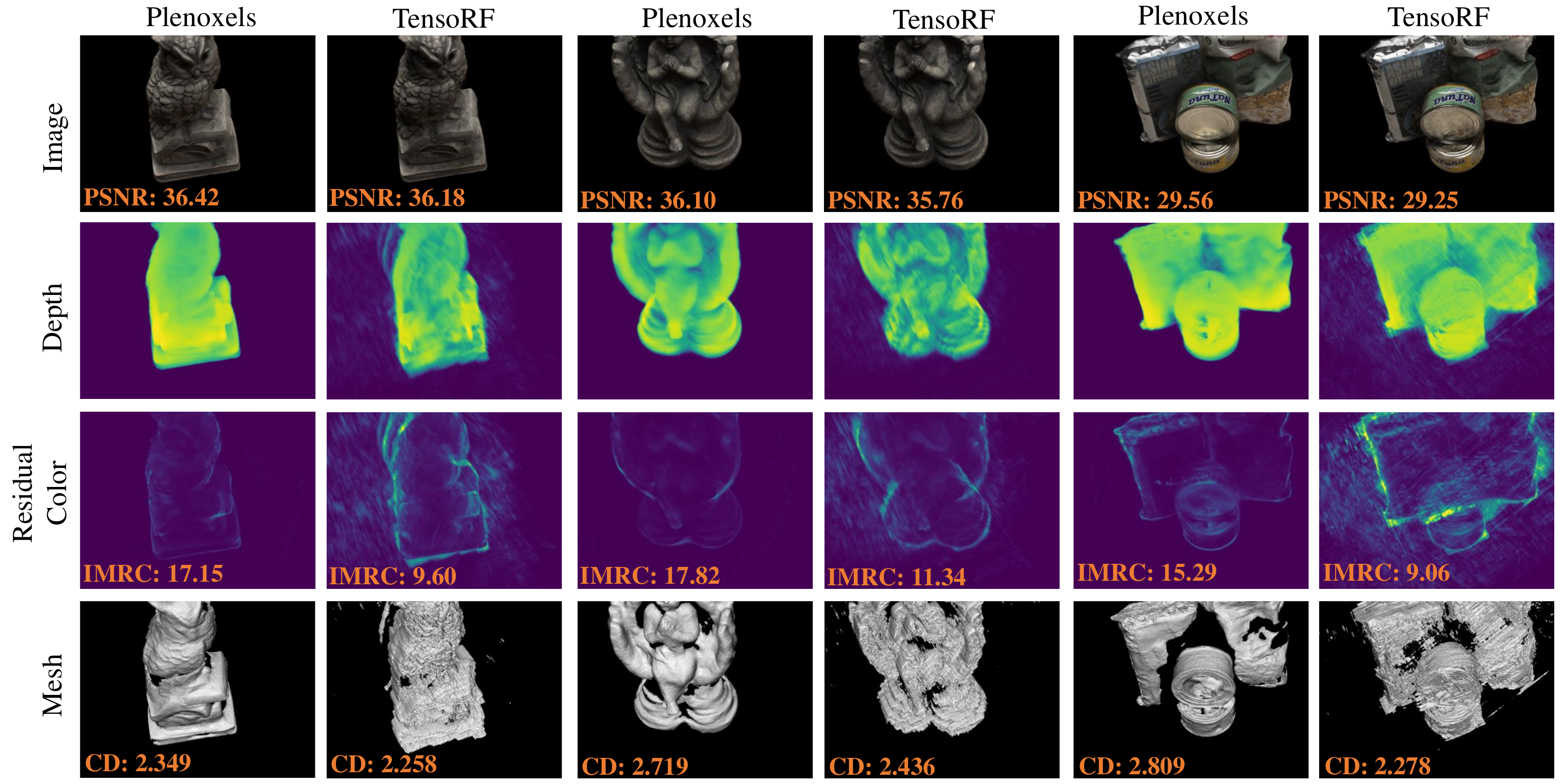}

   \caption{Three CD$\downarrow$/UserRank conflict cases. From left to right, DTU Scan 122, 118, and 97. For each scene, left has a better UserRank.}
   \label{fig:cd_user_2}
\end{figure}

\begin{figure}[ht]
  \centering
   \includegraphics[width=1\linewidth]{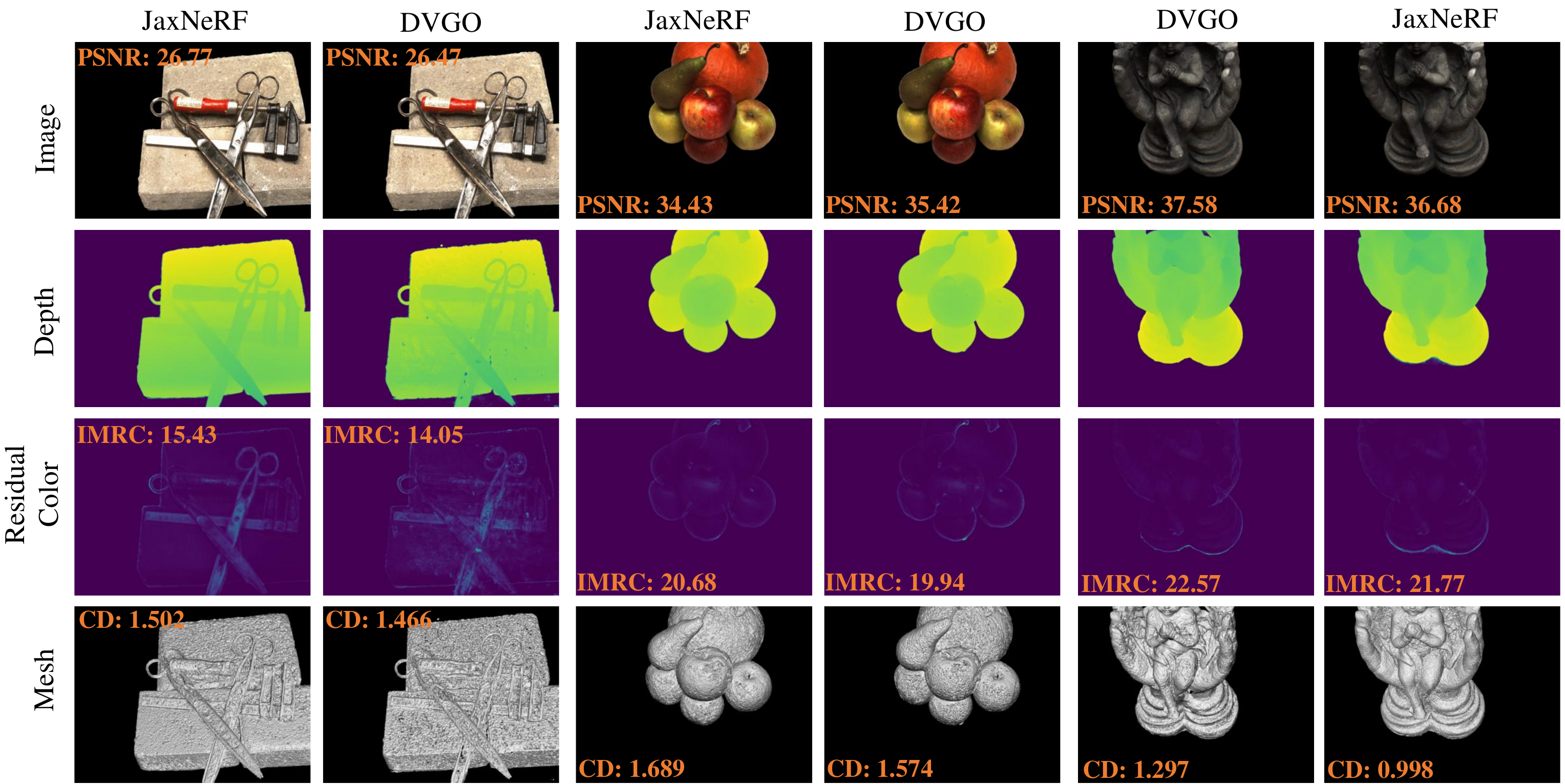}

   \caption{Three CD$\downarrow$/UserRank conflict cases. From left to right, DTU Scan 37, 63, and 118. For each scene, left has a better UserRank.}
   \label{fig:cd_user_3}
\end{figure}

We showcase $9$ remaining CD$\downarrow$/UserRank conflict cases in \cref{fig:cd_user_1} to \cref{fig:cd_user_3}. We have analysed the two main reasons that cause conflicts in Sec. 5.1. of the main paper. On one hand, because of the marching cubes algorithm, the CD metric fails to recognize some thick surfaces, and some low density surface points are discarded. On the other hand, the object mask used in the calculation neglects some floating meshes and leads to an unfair comparison.

\subsection{More CD/IMRC/UserRank Consistent Results}

\begin{figure}[ht]
  \centering
   \includegraphics[width=1.\linewidth]{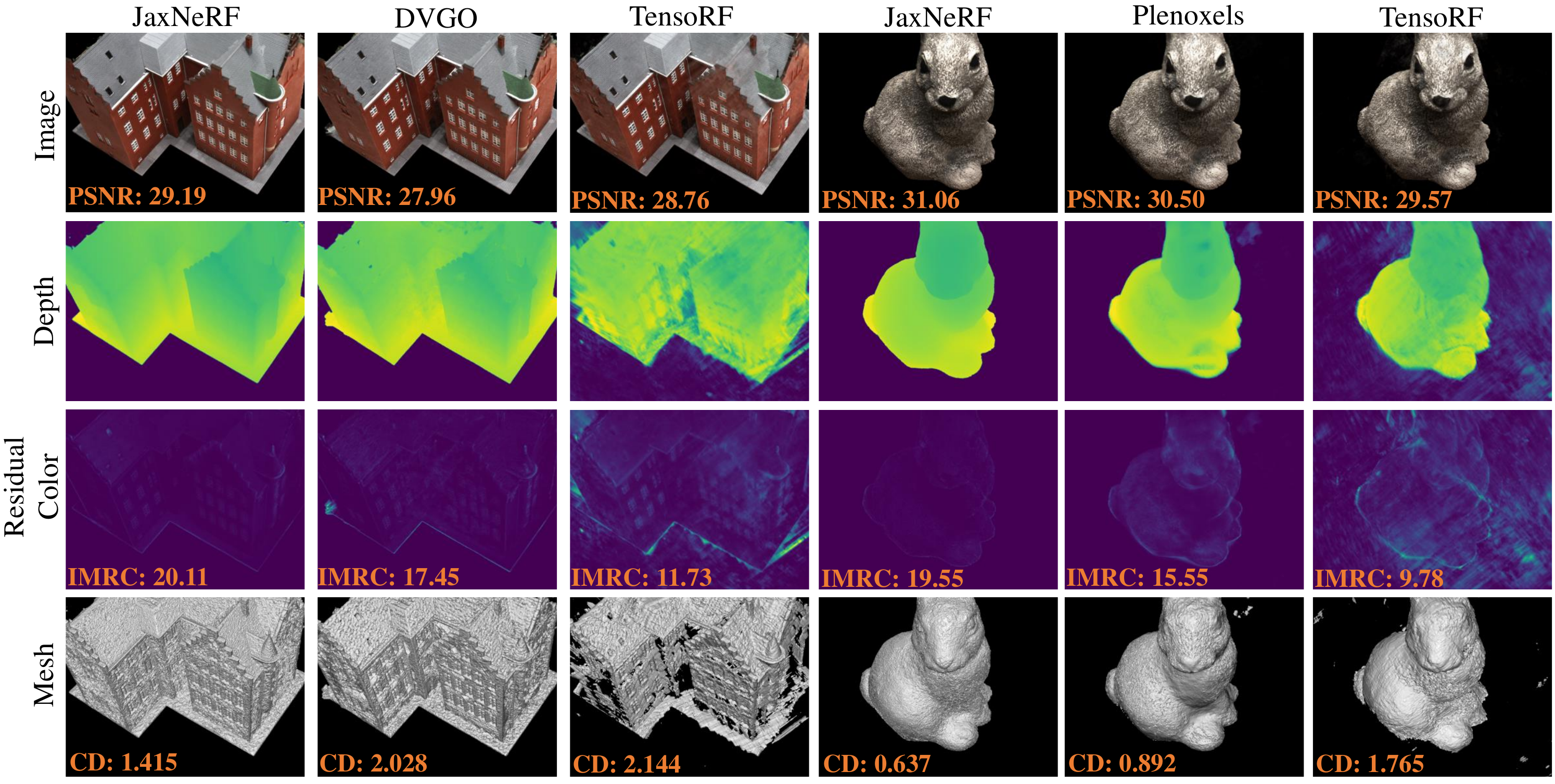}
   \caption{More CD$\downarrow$/IMRC$\uparrow$/UserRank consistent results on the DTU dataset. From left to right, Scan 24 and 55. For each scene, the quality of the density field decreases from left to right.}
   \label{fig:CD_IMRC_consistent1}
\end{figure}

\begin{figure}[ht]
  \centering
   \includegraphics[width=1.\linewidth]{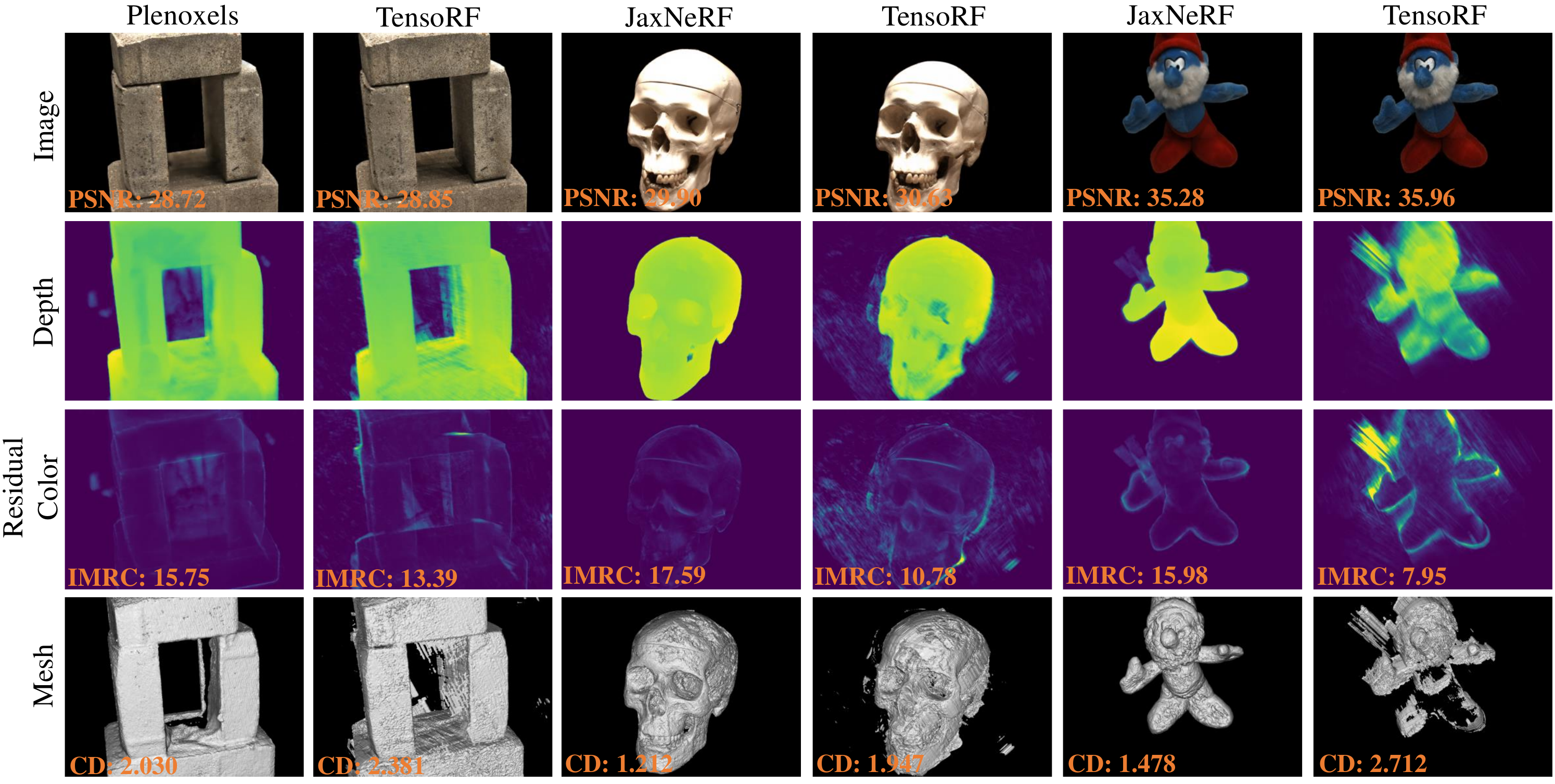}
   \caption{More CD$\downarrow$/IMRC$\uparrow$/UserRank consistent results on the DTU dataset. From left to right, Scan 40, 65, and 83. For each scene, the quality of the density field decreases from left to right.}
   \label{fig:CD_IMRC_consistent2}
\end{figure}

\begin{figure}[ht]
  \centering
   \includegraphics[width=1.\linewidth]{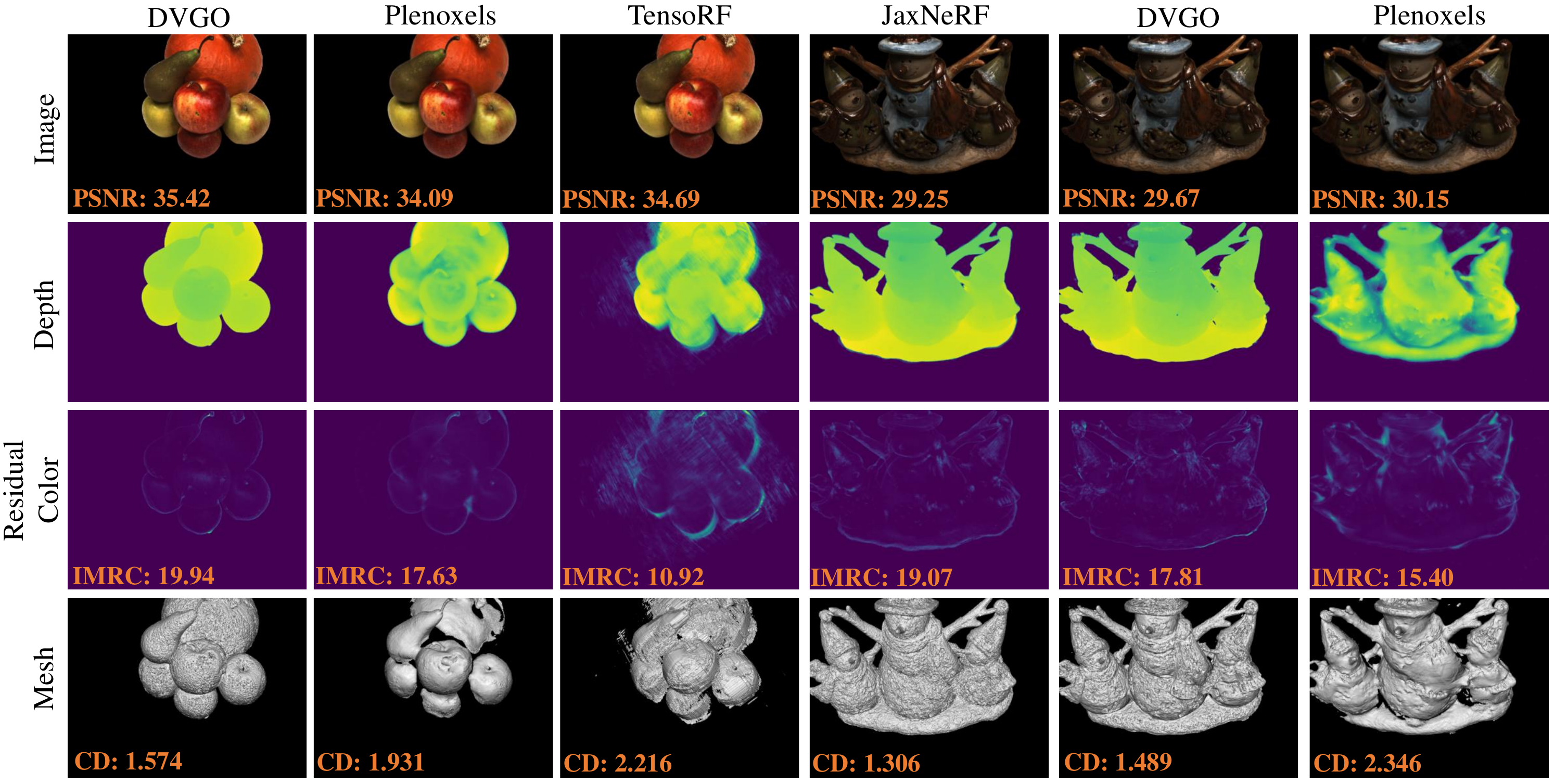}
   \caption{More CD$\downarrow$/IMRC$\uparrow$/UserRank consistent results on the DTU dataset. From left to right, Scan 63 and 69. For each scene, the quality of the density field decreases from left to right.}
   \label{fig:CD_IMRC_consistent3}
\end{figure}

\begin{figure}[ht]
  \centering
   \includegraphics[width=1.\linewidth]{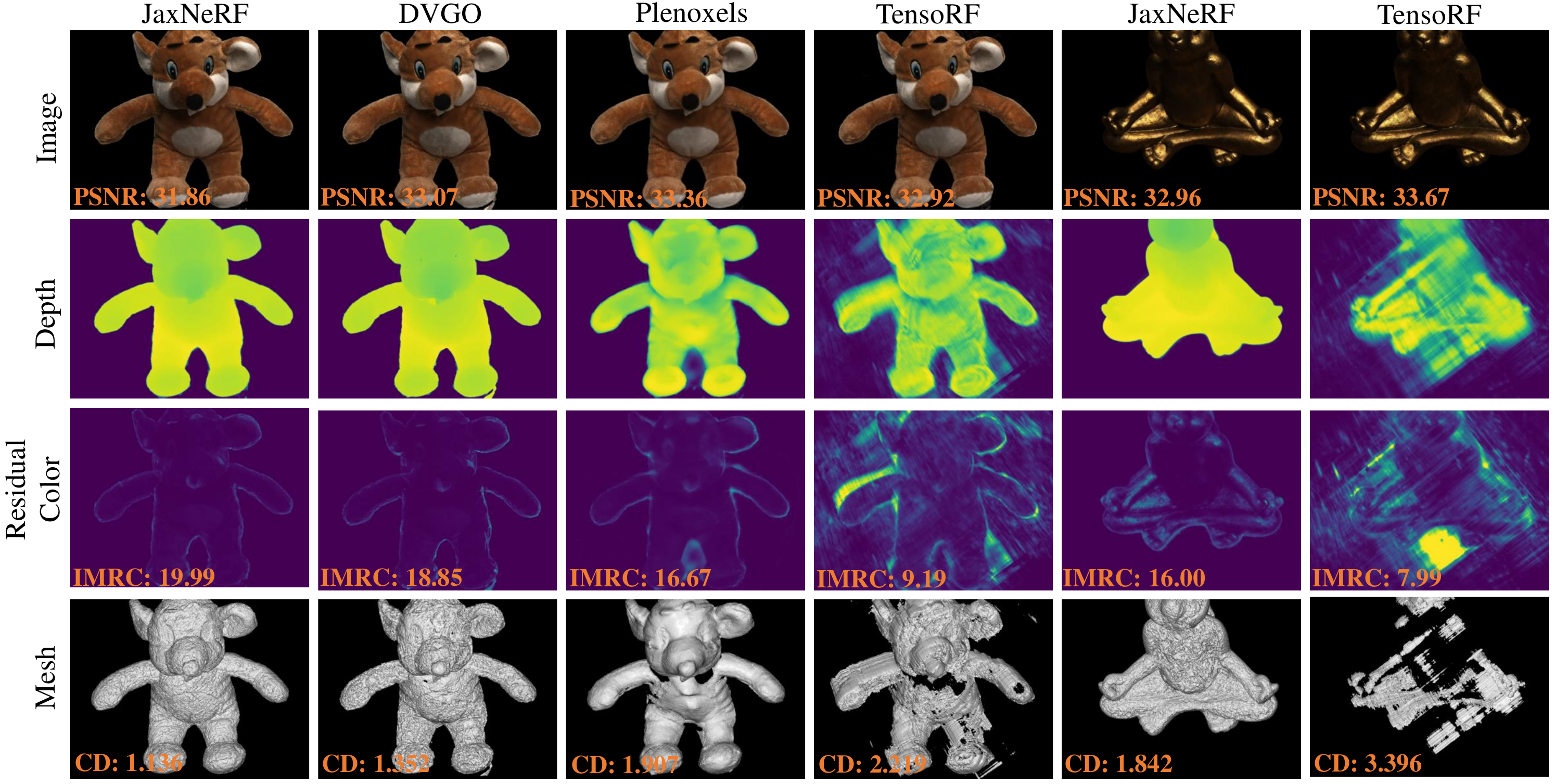}
   \caption{More CD$\downarrow$/IMRC$\uparrow$/UserRank consistent results on the DTU dataset. From left to right, Scan 105 and 110. For each scene, the quality of the density field decreases from left to right.}
   \label{fig:CD_IMRC_consistent4}
\end{figure}

\begin{figure}[ht]
  \centering
   \includegraphics[width=1.\linewidth]{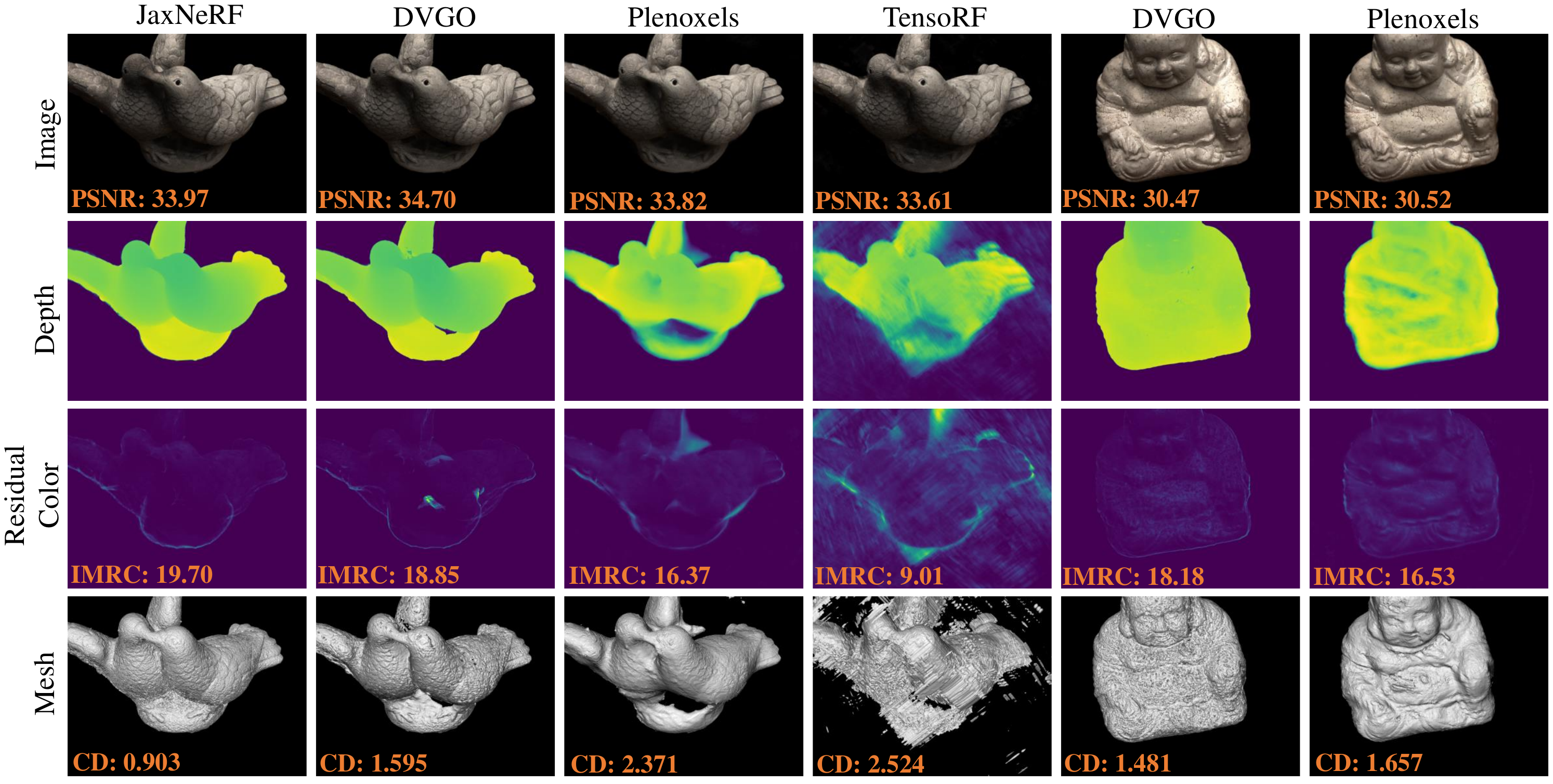}
   \caption{More CD$\downarrow$/IMRC$\uparrow$/UserRank consistent results on the DTU dataset. From left to right, Scan 106 and 114. For each scene, the quality of the density field decreases from left to right.}
   \label{fig:CD_IMRC_consistent5}
\end{figure}

On the DTU dataset, except for the $2$ IMRC$\uparrow$/UserRank and $11$ CD$\downarrow$/UserRank conflict cases, the remaining $77$ pairs all have consistent CD$\downarrow$/IMRC$\uparrow$/UserRank. We illustrate more consistent cases in \cref{fig:CD_IMRC_consistent1} to \cref{fig:CD_IMRC_consistent5}. We find that both CD and IMRC successfully reflect the quality of the density field in these cases. 

\end{document}